\documentclass[10pt,journal,compsoc]{IEEEtran}
\ifCLASSOPTIONcompsoc
\usepackage[nocompress]{cite}
\else
\usepackage{cite}
\fi
\ifCLASSINFOpdf
\else
\fi
\usepackage{makecell}
\usepackage{bbding}

\usepackage{graphicx}
\usepackage{subcaption}
\usepackage{float}
\usepackage{caption}
\usepackage{lscape}                                         
\usepackage{wrapfig}

\usepackage[lined,ruled,linesnumbered]{algorithm2e}
\usepackage{animate} 

\usepackage{booktabs}                   
\usepackage{multirow}
\usepackage{arydshln} %
\usepackage{enumitem}

\usepackage{bm}                          

\usepackage{dsserif}
\usepackage{amssymb}
\usepackage{amsmath}  

\usepackage{units}
\usepackage{color}
\usepackage[T1]{fontenc}    
\usepackage{amsfonts}       
\usepackage[utf8]{inputenc} 
\usepackage{nicefrac}       
\usepackage{microtype}      
\usepackage{pifont}         
\usepackage{comment}

\usepackage[mathscr]{euscript}
\usepackage{colortbl}

\usepackage{url}  

\usepackage[table]{xcolor}

\newlength\paramargin

\newcommand{\mfigure}[2]
{
\includegraphics[width=#1\linewidth]{#2}
}

\newcommand{\heading}[1]
{
\vspace{1mm}\noindent\textbf{#1}
}

\newcommand{\secref}[1]{Section~\ref{sec:#1}}
\newcommand{\figref}[1]{Figure~\ref{fig:#1}} 
\newcommand{\tabref}[1]{Table~\ref{tab:#1}}
\newcommand{\eqnref}[1]{Eq.~\eqref{eq:#1}}

\long\def\ignorethis#1{}

\newbox\jsavebox%

\makeatletter
\DeclareRobustCommand\onedot{\futurelet\@let@token\@onedot}
\def\@onedot{\ifx\@let@token.\else.\null\fi\xspace}

\def\eg{\emph{e.g}\onedot} 
\def\ie{\emph{i.e}\onedot}

\def\etal{\emph{et al}\onedot}
\makeatother

\graphicspath{{figures/}}
\usepackage{xcolor}
\colorlet{dark-blue}{blue!50!black}
\colorlet{dark-cyan}{cyan!75!black}
\colorlet{dark-purple}{purple!50!black}
\colorlet{dark-red}{red!75!black}
\colorlet{dark-green}{green!75!black}
\colorlet{dark-orange}{orange!50!black}
\colorlet{dark-gray}{black!75}
\colorlet{light-gray}{black!30}
\definecolor{nice-red}{HTML}{E41A1C}
\definecolor{nice-orange}{HTML}{FF7F00}
\definecolor{nice-yellow}{HTML}{FFC020}
\definecolor{nice-green}{HTML}{39b54a}
\definecolor{nice-blue}{HTML}{0071bc}
\definecolor{nice-purple}{HTML}{984EA3}

\colorlet{verylight-gray}{black!10}
\definecolor{LightCyan}{rgb}{0.88,1,1}

\definecolor{best}{rgb}{1, 0.85, 0.7}
\definecolor{second}{rgb}{1,1, 0.8}

\usepackage{ragged2e} 
\usepackage[pagebackref=true,breaklinks=true,letterpaper=true,colorlinks,bookmarks=false,urlcolor=blue,citecolor=blue]{hyperref}

\hypersetup{
	citecolor=dark-green,
	filecolor=blue,
	linkcolor=nice-red,
	urlcolor=nice-blue 
}

\begin{document}
\bstctlcite{IEEEexample:BSTcontrol}
	%
	\title{Moment-Reenacting: Inverse Motion Degradation with Cross-shutter Guidance}
	\author{Xiang~Ji, Guixu~Lin, Zhengwei Yin, Jiancheng Zhao,  and Yinqiang Zheng
		\IEEEcompsocitemizethanks{
            \IEEEcompsocthanksitem Xiang Ji, Guixu Lin, Zhengwei Yin, Jiancheng Zhao and Yinqiang Zheng are with the Graduate School of Information Science and Technology, The University of Tokyo, Japan.  \protect
			\IEEEcompsocthanksitem Yinqiang Zheng is the corresponding author. E-mail: yqzheng@ai.u-tokyo.ac.jp \protect
		}
	}

	\IEEEtitleabstractindextext{%
		\justify
\begin{abstract}
Motion degradation, manifested as blur in global shutter (GS) images or rolling shutter (RS) distortion in RS counterparts, remains a fundamental challenge in computational imaging, especially under fast motion or low-light conditions. While prior works have treated blur decomposition and RS temporal super-resolution as separate tasks, this separation fails to exploit their intrinsic complementarity. In this paper, we propose a unified framework to invert motion degradation and reenact imaging moment by jointly leveraging the complementary characteristics of GS blur and RS distortion. To this end, we introduce a novel dual-shutter setup that captures synchronized blur-RS image pairs and demonstrate that this combination effectively resolves temporal and spatial ambiguities inherent in both modalities. For allowing flexible performance-cost trade-offs, we further extend this dual-shutter setup to a stereo Blur-RS configuration with a narrow baseline. In addition, we construct a triaxial imaging system to collect a real-world dataset with aligned GS-RS pairs and ground-truth high-speed frames, enabling robust training and evaluation beyond synthetic data. Our proposed network explicitly disentangles motion into context-aware and temporally-sensitive representations via a dual-stream motion interpretation module, followed by a self-prompted frame reconstruction stage. Extensive experiments validate the superiority and generalizability of our approach, establishing a new paradigm for realistic high-speed video reconstruction under complex motion degradations. Codes and more resources are available at \url{https://jixiang2016.github.io/dualBR_site/}.
\end{abstract}

		}

	\maketitle

	\IEEEdisplaynontitleabstractindextext

	%
	\IEEEpeerreviewmaketitle


	%
	%
	%
	%

\IEEEraisesectionheading{
\section{Introduction}
\label{sec:intro}}

\begin{figure}[!tb]
	\centering
	{
			\mfigure{1}{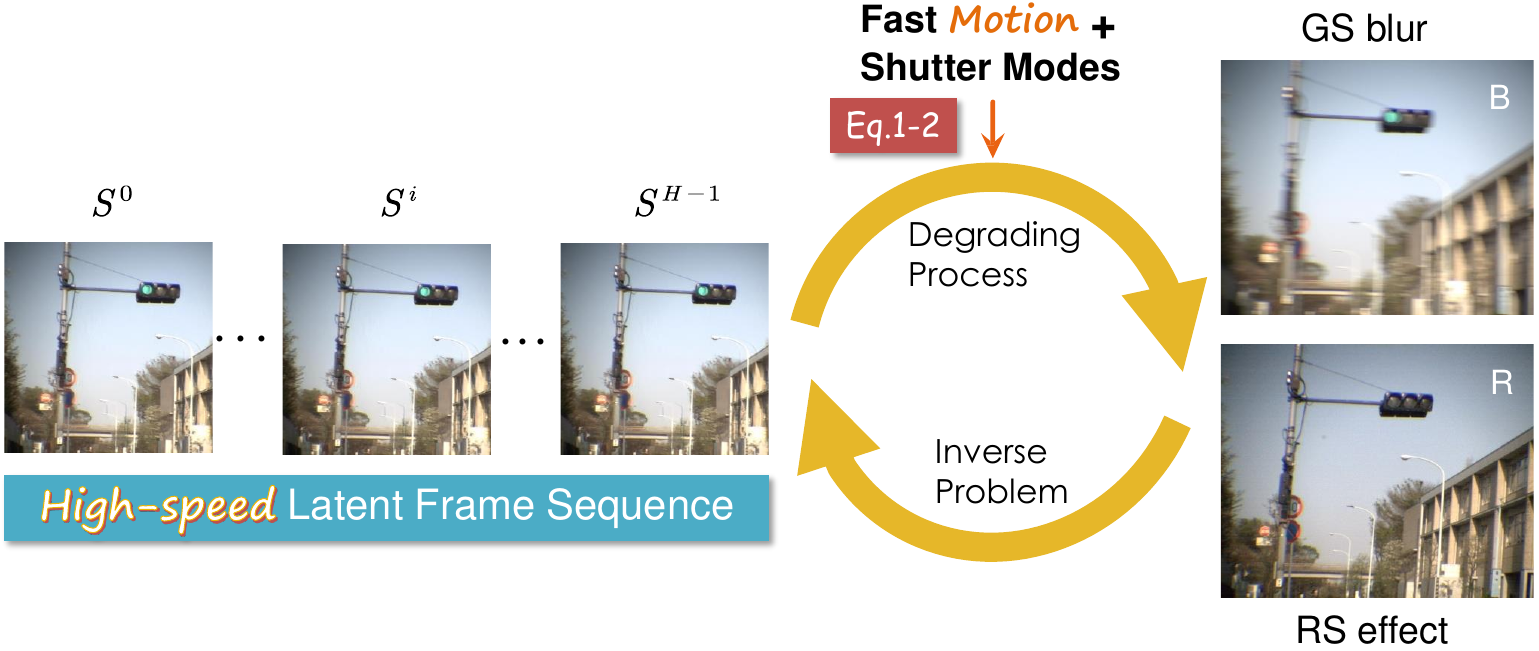}
	}
	\vspace{-5mm}
	\caption{
		\textbf{Imaging moment reenacting.} If latent sharp frames within an exposure captured by GS or RS modes under fast motion, it will cause distinct degradations: blur and RS effects. Current research reverses the two degrading processes independently. We propose to simultaneously handle the two issues by exploiting the complementarity of RS and GS views, and ultimately reenact the imaging moment regardless of motion degradations.
	}
	\label{fig:moment_reenacting}
	\vspace{-4mm}
\end{figure}

\IEEEPARstart{P}{hotographic} capture relies on sufficient exposure time to accumulate photons. When fast motions get involved, existing imaging system will inevitably introduce undesired degradations, motion blur or rolling shutter effect, exhibiting significant relevance to the exposure modes exploited: Global Shutter (GS) and Rolling Shutter (RS) respectively (\figref{moment_reenacting}). These motion degradations severely compromise perceptual quality and content interpretability. Hence, extensive research has been devoted to reverse these degrading processes and produce sharper details.

While conventional approaches formulate reconstruction as single-frame recovery through image-to-image translation (\ie, deblurring~\cite{nah2017deep,tao2018scale,wang2022uformer,gao2019dynamic,zamir2022restormer,purohit2020region,zhou2019davanet,ren2023multiscale,ji2023single,wang2022neural} or RS correction~\cite{rengarajan2016bows,purkait2017rolling,forssen2010rectifying,grundmann2012calibration,vasu2018occlusion,zhuang2017rolling,zhuang2019learning,yan2023deep,rengarajan2017unrolling,fan2023joint}). Recent advances have shifted focus toward more ambitious tasks, retrieving a sharp image sequence instead, dubbed as blur decomposition~\cite{jin2018learning,pan2019bringing,purohit2019bringing,zhong2022blur} and RS temporal super-resolution~\cite{fan2021inverting,zhong2022bringing,shang2023self,fan2022context}. Despite significant progress, both tasks have evolved in a separate manner, with no rigorous studies exploring the potential of their correlations for the inversion process. Moreover, they are basically grounded on synthetic data~\cite{nah2017deep,shen2020video,liu2020deep,zhong2022bringing}, leaving the real-world performance uncertain. Such synthesis processes can easily lead to unnatural artifacts~\cite{rim2020real,zhong2021towards,ji2023rethinking}. Thus, models trained on synthetic datasets are likely to demonstrate inferior generalization and may not perform effectively in practice.

\begin{figure}[!tb]
	\centering
	{
			\mfigure{1}{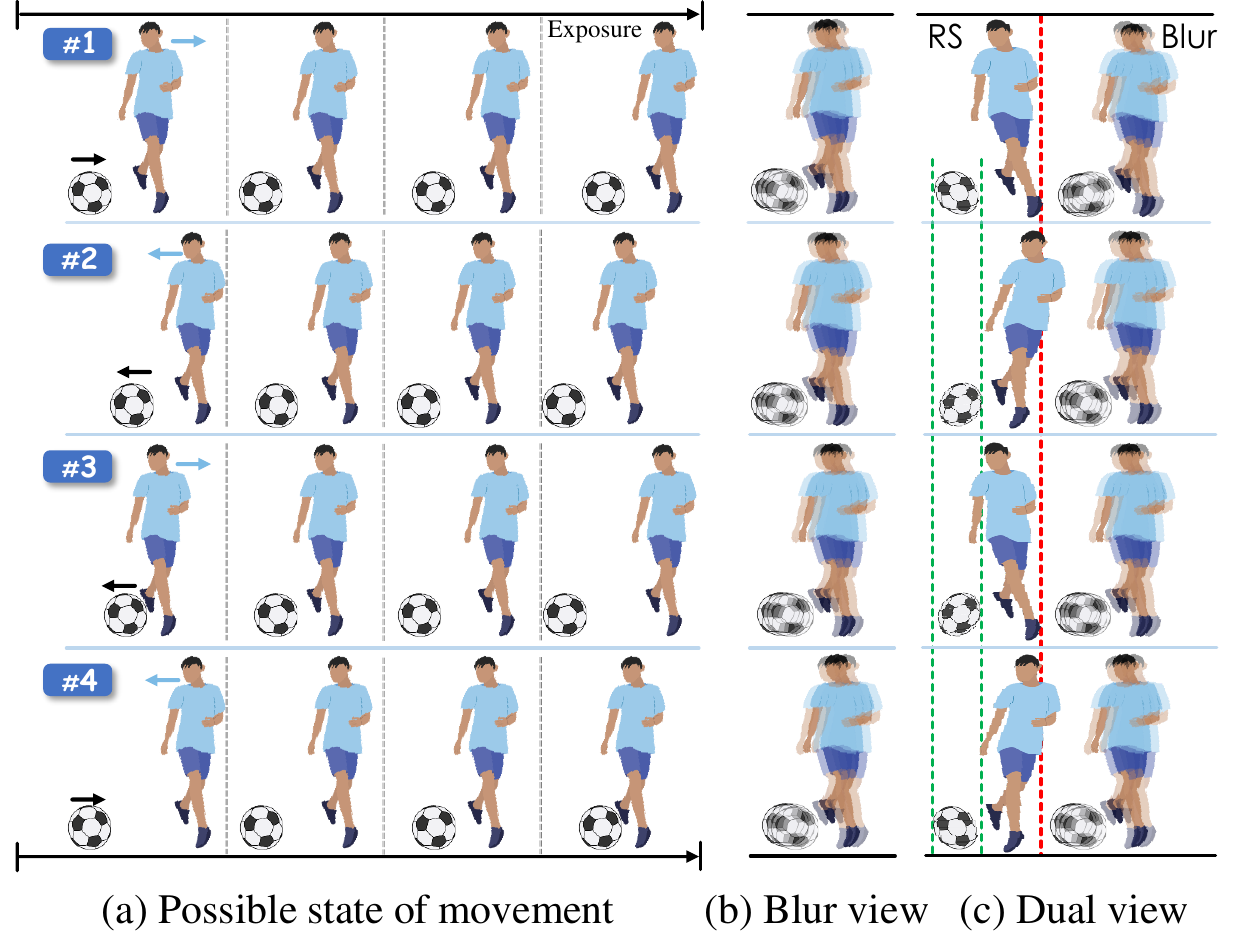}
	}
	\vspace{-3mm}
	\caption{
		\textbf{Motion ambiguity of blur observation.} In this toy example, we show two objects: a soccer and a player, both moving horizontally. (a) shows four possible motion states (both moving right, both moving left, one moving left while the other moves right.) during the exposure time. (b) presents corresponding motion blurred observations. 
		They are all identical due to averaging effects, which brings about motion ambiguity to blur decomposition and further causes issues of instability and low performance~\cite{zhong2022animation}. (c) In our dual Blur-RS setting, rolling shutter (RS) view implicitly encodes temporal ordering of latent frames.
	}
	\label{fig:motion_ambiguity}
	\vspace{-2mm}
\end{figure}

\begin{figure}[!t]
	\centering
	{
		\mfigure{1}{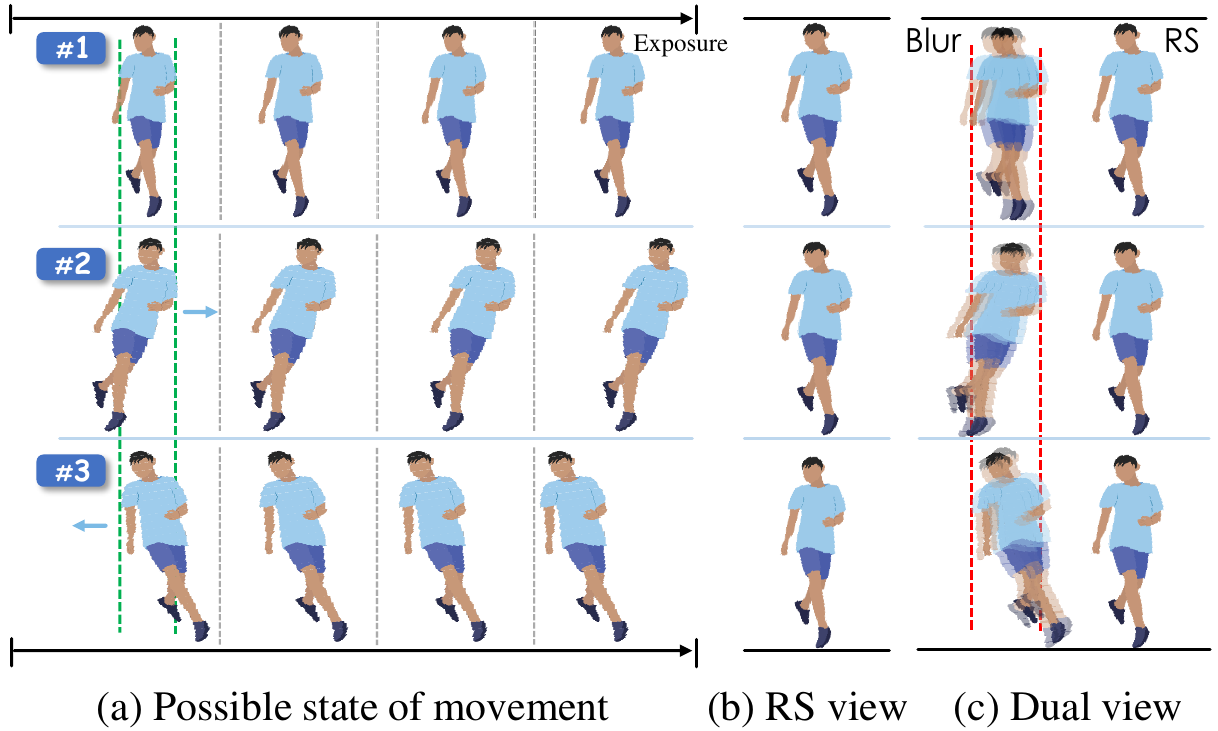}
	}
	\vspace{-3mm}
	\caption{
		\textbf{Initial-state ambiguity of RS observation.} We show a man moving horizontally with different poses (upright, tilted to the right, and tilted to the left). (a) shows three possible motion states (static, moving right and moving left) during the exposure time. (b) presents corresponding RS observations. They are all identical due to different initial-state, which brings about ambiguity to RS temporal super-resolution. (c) In our dual Blur-RS setting, blur view sufficiently indicates initial-state of latent frames.
	}
	\label{fig:initial_state_ambiguity}
	\vspace{-4mm}
\end{figure}

Most importantly, the intrinsic ambiguities in the degradation processes make both problems even more difficult. Under the context of blur decomposition, averaging effects of motion blur have severely destroyed the temporal ordering of latent frames~\cite{jin2018learning}. Moreover, the presence of motion ambiguity within each dynamic object intensifies the problem, generating numerous plausible yet physically unrealistic solutions. This ambiguity is illustrated in ~\figref{motion_ambiguity}a-b with two dynamic objects. Successfully resolving the loss of temporal order in blur decomposition remains an ongoing challenge. Existing approaches can be broadly classified into two types: (i) methods that employ ordering-invariant loss~\cite{jin2018learning} and (ii) techniques that approximate latent temporal order within exposure by leveraging the motion of consecutive blurred frames~\cite{zhong2022animation,zhong2022blur}. The first category often leads to sub-optimal solutions due to the limited supervision provided by the loss function, whereas the second struggles with significant degradation under conditions of prolonged exposure or rapid motion. Additionally, estimating motion within blurry frames is inherently complex and computationally expensive. RS images have recently been acknowledged for encoding hidden motion due to their row-by-row exposure mode~\cite{fan2021inverting,fan2022context}, which provides a foundation for mitigating RS effects. However, reconstructing a sequence from a single RS image remains unresolved due to the absence of complete global content. Conversely, blurred images preserve globally integrated scene information but lack temporal order. While human visual perception is less sensitive to RS effects compared to blur, the severe ill-posedness of RS temporal super-resolution makes it not only nontrivial but even more difficult than blur decomposition~\cite{Ji_2023_ICCV}. One potential reason is the ambiguity in the initial state, as illustrated in ~\figref{initial_state_ambiguity}a-b. For instance, when an RS image captures an upright figure, the unknown initial state leads to three possible motion patterns within the exposure period.

Therefore, considering the complementarity of Blur and RS observations, we propose dual Blur-RS setting to reenact the imaging moment by jointly handling the aforementioned ambiguities of blur decomposition and RS temporal super-resolution. As shown in ~\figref{motion_ambiguity}c, the RS view not only provides local details but also implicitly captures temporal order of latent frames. Meanwhile, the GS view could be exploited to disambiguate the initial-state from RS counterpart (\figref{initial_state_ambiguity}c). On the other hand, dual camera system has been widely exploited in RS correction (RS-RS, RS-Event)~\cite{albl2020two,zhong2022bringing,zhou2022evunroll,lin2023event}, deblurring (GS-Event) ~\cite{sun2022event,xu2021motion}, and even vibration sensing (GS-RS)~\cite{sheinin2022dual}, which offers a promising pathway to address synthetic data challenges. Inspired by the hardware design of ~\cite{rim2020real,zhong2021towards}, we devise our triaxial imaging system to capture strictly aligned high-speed sharp videos and low-speed Blur-RS pair videos. Facilitated by the imaging prototype, we contribute a real world dataset (realBR) capable of benchmarking multiple correlated tasks (\eg RS correction, deblurring and video frame interpolation).

Beyond the collected dataset, we further introduce a novel model to recover a sharp video sequence from cross-shutter views, specifically blur and RS observations, through two stages: motion interpretation and frame reconstruction. The motion interpretation module first decomposes the bilateral motion fields into complementary streams: Blur and RS branches, where each branch highlights distinct aspects, contextual characterization and temporal abstraction. Shutter alignment and aggregation facilitate the mutual boosting of both branches by enabling the interaction of aligned and aggregated features. Furthermore, incorporating temporal positional encoding for the RS view improves the model’s capacity to resolve uncertainties in motion direction. Subsequently, estimated motion fields along with blur-RS inputs will be warped and refined through frame reconstruction module, self-prompted by motion residues, to generate a sharp video clip. To compensate for the absence of motion supervision, we introduce adaptive task-oriented distillation combined with reconstruction loss, further regulating and solidifying the training phase.

In summary, our main contributions are:
\begin{itemize}
	\item \textbf{Problem:} We present, for the first time, a new setting of dual Blur-RS combination to address the ambiguities of blur decomposition and RS temporal super-resolution, demonstrating its superiority over pure RS/blur and existing cross-sensor settings. Then we further extend the dual set-up to a more frequent one: stereo Blur-RS with a narrow baseline.

    \item \textbf{Data:} Facilitated by beam splitters, we develop a triaxial imaging system that captures spatially and temporally aligned Blur-RS pairs along with high-speed ground truth. The collected real dataset contains various ego-motion and object-motion types, effectively superseding traditional numerical simulation.

    \item \textbf{Algorithm:} We propose a novel neural network architecture that actively addresses contextual characterization and temporal abstraction through a dual-stream motion interpretation module, augmented by region-adaptive distillation. And our self-prompted frame reconstruction can dynamically identify and refine discrepancies between bilaterally warped frames.
    
    \item  \textbf{Evaluation:} Extensive comparisons with SOTA methods thoroughly validate the effectiveness of our setting and model. Furthermore, rigorous cross-validation protocols, independent third-party testing, and downstream task evaluations collectively demonstrate the superiority, generalizability and practicality of real data-trained models over synthetic data-based counterparts.
\end{itemize}

A preliminary version of this work was published as a conference paper~\cite{ji2024motion}. In this paper, the main extensions can be summarized as: (1) We shift our focus from original blur decomposition to a more general problem, motion degradation inversion, which addresses the RS temporal super-resolution and blur decomposition equally. (2) The motion interpretation and frame reconstruction modules are further enhanced respectively, by incorporating region-adaptive distillation to regularize intermediate flow learning and motion-residue prompt to accurately guide refinement's focus towards critical areas. (3) We address the referred limitation in the preliminary version by extending the dual set-up to stereo one with narrow baseline, enabling performance-cost trade-offs according to specific needs. (4) We enrich our test set by collecting third-party data with different capturing parameters and new stereo data with various baselines. The real data imperatives and synthetic data constraints are also comprehensively examined through cross-validation, third-party testing and downstream applications.

	\vspace{-2mm}
\section{Related Work}
\label{sec:relate}

\subsection{Blur Decomposition}  

Reconstructing an image sequence from a single blurred input is more challenging than conventional deblurring, as temporal ordering of latent frames is lost due to exposure-time averaging. Jin et al.~\cite{jin2018learning} first identify this ambiguity and mitigate it using an ordering-invariant loss. Purohit et al.~\cite{purohit2019bringing} learn motion representations from sharp videos via a surrogate task and use them to guide motion encoding for blurred inputs, while Argaw et al.~\cite{argaw2021restoration} adopt an end-to-end framework with spatial transformer modules to jointly recover video frames and motion. To address directional ambiguity, BiT~\cite{zhong2022blur} exploits three consecutive blurry frames to extract motion priors and recovers arbitrary sharp frames through blur intra-interpolation. Zhong et al.~\cite{zhong2022animation} further introduce multiform motion guidance, obtained via learning, approximation from blurry videos, or user input, though learning-based guidance still relies on flow supervision from consecutive blurry frames. Overall, these methods~\cite{purohit2019bringing,zhong2022blur,zhong2022animation} share a common strategy of leveraging multiple blurry inputs to model latent motion.
 
Unlike blur decomposition, another approaches build upon the framework of video frame interpolation by using blurry frames as input to restore clear images within the deadtime. Most methods in this category process a sequence of images and generate a sharp frame positioned between two blurry ones~\cite{shen2020blurry,zhang2020video,jin2019learning}. More recent studies have extended this concept to synthesize interpolated frames at arbitrary time points~\cite{oh2022demfi,ji2023rethinking}. In this paper, our primary focus is on the motion ambiguity in blur decomposition, so we do not delve into these works in detail.

\subsection{RS Temporal Super-Resolution}
Extracting high-frame-rate GS videos from RS-degraded observations has attracted growing interest, yet remains highly ill-posed and inherently ambiguous. Early approaches reduce ambiguity via explicit motion assumptions, such as constant velocity~\cite{zhuang2017rolling,fan2021inverting} or acceleration~\cite{fan2022rolling}, which often fail to hold under complex real-world motions. Recent studies therefore favor learning-based approaches that exploit consecutive RS frames~\cite{fan2022context,fan2024learning,fan2024unified}. Specifically, Fan et al.~\cite{fan2022context} propose CVR, a context-aware video reconstruction network for occlusion reasoning and motion compensation. They further introduce a unified framework with shared parameters to support multiple shutter types~\cite{fan2024unified}. Moreover, LBCNet~\cite{fan2024learning} models uniform bilateral motion fields with time-offset encoding to enable rolling-shutter-aware backward warping and enforce spatio-temporal consistency.

Beyond using two adjacent RS frames, Zhong \emph{et al}\onedot~\cite{zhong2022bringing} introduce dual reversed RS inputs to better handle complex motion and occlusions. Additional techniques, including self-supervision\cite{shang2023self,fan2024self} and event stream processing~\cite{zhou2022evunroll,lu2024uniinr}, further enhance RS inversion. Instead of applying cycle consistency on dual-reversed RS images~\cite{shang2023self}, Fan \emph{et al}\onedot~\cite{fan2024self} develop a conversion model of RS2GS and GS2RS for self-supervised learning. While event cameras, with their high temporal resolution, enable the reconstruction of sharp GS frames from RS inputs by bridging spatio-temporal gaps~\cite{zhou2022evunroll}. Despite these advancements in alleviating ambiguity, most existing methods are trained and evaluated primarily on synthetic data, and their robustness remains limited in highly dynamic real-world scenes.

\vspace{-1mm}
\subsection{Dual Camera System}
Recently, notable progress has been achieved by utilizing dual-camera systems for various vision tasks. According to their sensor combinations, these systems can be broadly categorized into RS-RS~\cite{albl2020two,zhong2022bringing}, RS-Event~\cite{zhou2022evunroll,lu2024uniinr}, GS-Event~\cite{sun2022event,xu2021motion,weng2023event}, and RS-GS~\cite{sheinin2022dual} setups.

Addressing the ill-posed nature of single-image RS correction, Albl \emph{et al}\onedot~\cite{albl2020two} propose a dual-RS setup with opposite scanning directions and show that it enables RS effects correction from sparse point correspondences. Zhong \emph{et al}\onedot~\cite{zhong2022bringing} later extend this idea with an end-to-end learning framework that iteratively predicts dual optical-flow sequences. Beyond RS-RS systems, Zhou \emph{et al}\onedot~\cite{zhou2022evunroll} pair an RS sensor with an event camera to mitigate RS distortions, while prior GS-Event methods~\cite{sun2022event,xu2021motion} exploit events for motion deblurring: Sun \emph{et al}\onedot~\cite{sun2022event} integrate event and image features via cross-modal attention, and Xu \emph{et al}\onedot~\cite{xu2021motion} address modality inconsistency using a piecewise linear motion model for self-supervised deblurring. Additionally, Sheinin \emph{et al}\onedot~\cite{sheinin2022dual} use a mixed RS/GS dual-camera system to capture and compensate for vibrations.

Overall, these studies have demonstrated the feasibility of dual-camera systems in improving accuracy, robustness, and adaptability under complex visual scenarios, providing a solid foundation for the approach explored in this paper. Importantly, prior work~\cite{zhong2022bringing} indicates that such systems remain effective under imperfect cross-camera alignment and do not strictly rely on fully synchronized or perfectly matched observations, which in turn motivates the proposed dual Blur–RS setting and the accompanying robustness analysis.

\section{Imaging Moment Reenacting}
\label{sec:method}

\begin{figure}[!t]
	\centering
	{
		\mfigure{1}{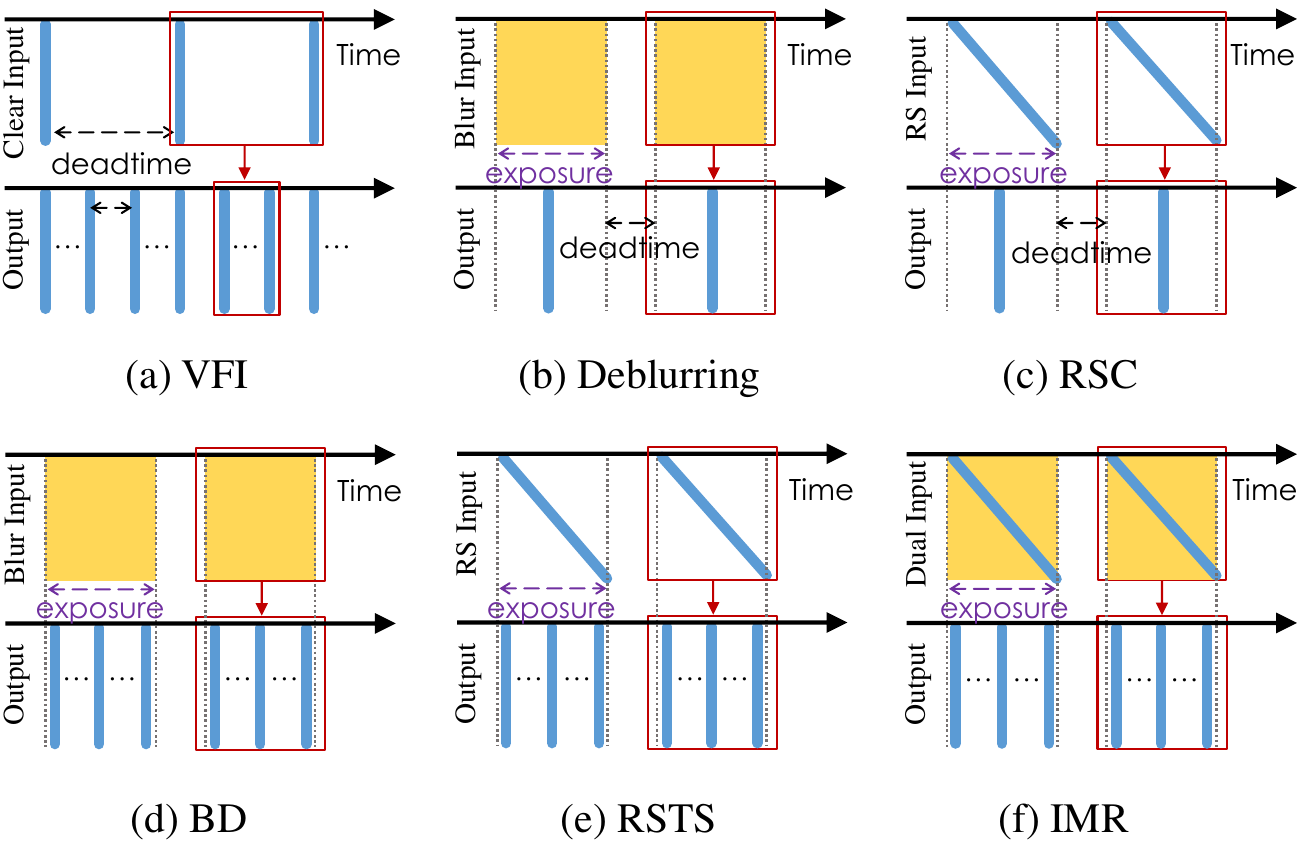}
	}
	\vspace{-5mm}
	\caption{
		\textbf{Comparisons of related tasks.} The horizontal and vertical dimension denote time and image rows. (a) Video Frame Interpolation (VFI). (b) Deblurring. (c) RS Correction (RSC). (d) Blur Decomposition (BD). (e) RS Temporal Super-resolution (RSTS). (f) Our proposed Imaging Moment Reenacting task(IMR).
	}
	\label{fig:formulation}
	\vspace{-2mm}
\end{figure}

\subsection{Formulation and Motivation}
\label{sec:setup}

Conventional approaches model blur as the convolution of sharp images with spatially varying kernels~\cite{hirsch2011fast,harmeling2010space}. However, such kernel-based approximations are often inaccurate, especially under abrupt motion discontinuities and occlusions~\cite{nah2017deep}. To better mimic the physical imaging process, prior works~\cite{nah2017deep,oh2022demfi,zhong2022blur,ji2023single} formulate a blurred frame $B$ as the temporal integration of latent frames during exposure:
\begin{equation}
	\label{eq:blur}
	\begin{split}
		B = g\left(\frac{1}{T}\int_{t=0}^{T}A^t dt\right) \simeq \frac{1}{L} \sum_{t=0}^{L-1} S^t ,
	\end{split}
\end{equation}
where $T$ and $A^t$ denote the exposure time and sensor signals in radiance space at time $t$, and $g$ is the camera response function (CRF). In practice, this formulation is discretized as averaging $L$ sampled frames in the sRGB domain.
In contrast, RS cameras acquire images by exposing pixels sequentially, row by row. Therefore, RS distortions (also known as jello effects) arise when motion occurs during acquisition, and can be formulated as~\cite{liu2020deep,ji2023rethinking,fan2021inverting}:
\begin{equation}
	\label{eq:rs}
	\begin{split}
		[R]_t = [S^t]_t, \quad t = 0,1,\cdots,L-1 ,
	\end{split}
\end{equation}
where $[S^t]_t$ denotes extracting the $t$-th row from the latent sharp frame $S^t$ captured at time $t$, and $L$ equals the height of the RS image.

These degradations reduce the interpretability of visual content, motivating extensive research on recovering clear details. Generally, deblurring (\figref{formulation}b) and RS correction (\figref{formulation}c) techniques address this by transforming one degraded image into a sharper version, typically by estimating one of latent frames within the exposure period. More recently, the focus has shifted toward reversing the degradation process entirely to reconstruct a sequence of latent frames through blur decomposition (\figref{formulation}d) or RS temporal super-resolution (\figref{formulation}e).
Unlike video frame interpolation task (\figref{formulation}a), which interpolates frames in deadtime between adjacent sharp inputs, these tasks aim to recover latent frames within the exposure time. Although significant progress has been made, blur decomposition and RS temporal super-resolution have developed independently, and no thorough studies have investigated the potential correlations between them for improving the inversion process. As we discussed in \secref{intro}, both tasks are highly ill-posed, yet exhibit a large degree of complementarity. RS view not only provides local details but also implicitly captures temporal order of latent frames. Meanwhile, blur observation could be exploited to mitigate the initial-state ambiguity from RS counterpart. Therefore, we propose to simultaneously handle the two issues by recovering the latent frame sequence, and ultimately reenact the imaging moment (\figref{formulation}f) regardless of motion degradations. 

\begin{figure*}[!tb]
	\centering
	\mfigure{1}{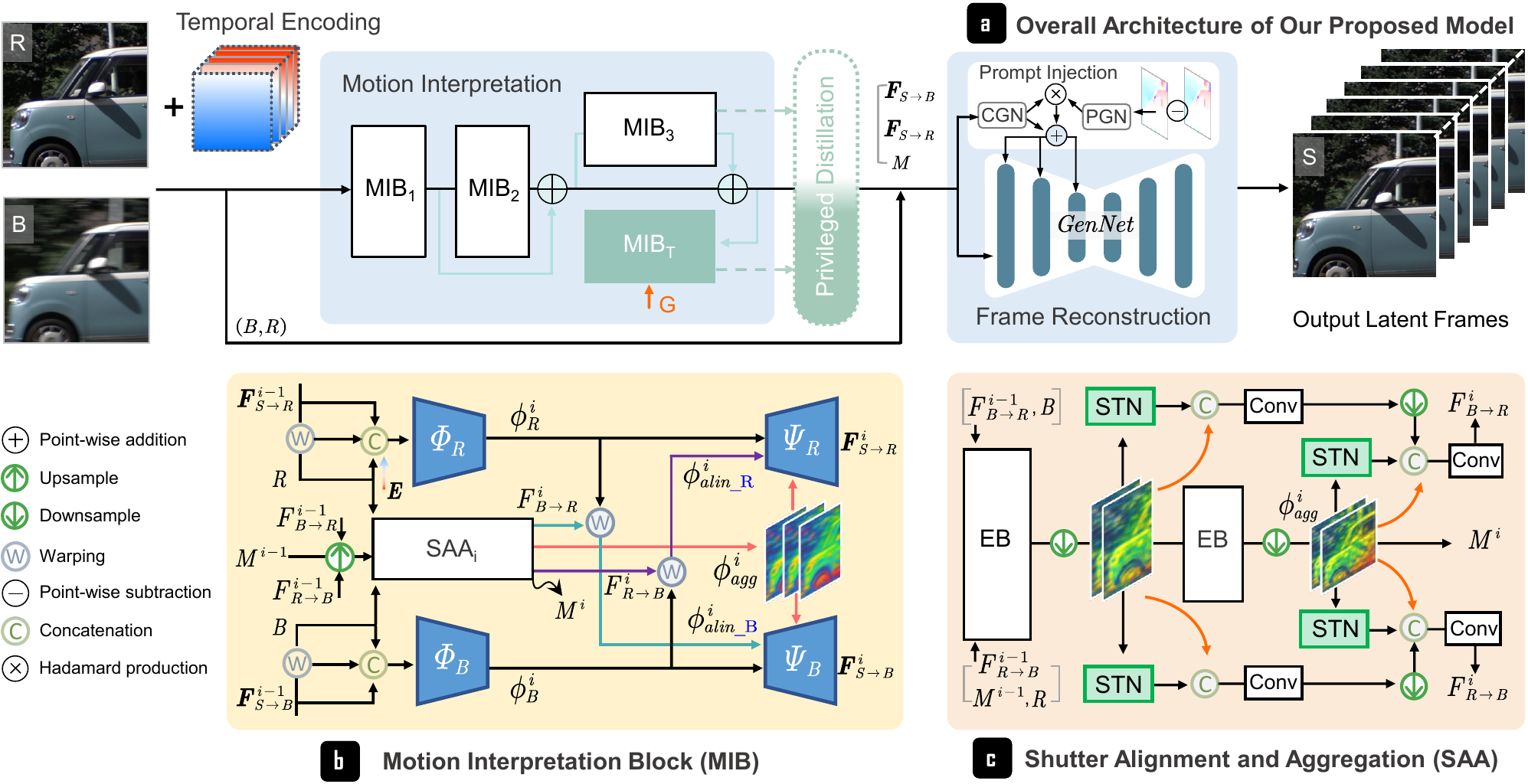}
	\vspace{-4mm}
	\caption{
		\textbf{Our proposed model.} (a) illustrates the overall architecture, which comprises two main stages: motion interpretation and frame reconstruction. The motion interpretation takes as input a blurry image $B$, a RS image $R$, and its corresponding temporal positional encoding $\textbf{E}$. It includes three student blocks MIB$_i$ and one teacher MIB$_T$. They follow the same structure as in (b), except that the teacher module is additionally provided with the ground-truth $\textbf{G}$. (c) elaborates on the Shutter Alignment and Aggregation (SAA) component, where a Spatial Transformer Network (STN) is introduced to estimate a global transformation for coarse feature alignment between the blur and RS streams, complementing the flow-based local warping. The frame reconstruction is implemented through a \emph{GenNet}, incorporating both Prompt Generation Network (PGN) and Context Generation Network (CGN) to generate the predicted latent frames $\textbf{S}$.
	}
	\label{fig:model}
	\vspace{-3mm}
\end{figure*}

\subsection{Network Architecture}
\label{sec:model}

The overview of our proposed architecture is shown in \figref{model}a. In general, the model framework is structured into two sequential stages: motion interpretation followed by frame reconstruction. The motion interpretation (MI) phase is designed to iteratively exploit the merits of our Blur-RS combination, employing three motion interpretation blocks ($MIB_1$, $MIB_2$, $MIB_3$) under the guidance of a teacher module ($MIB_T$). 
The estimation of latent motion fields can be described as below:
\begin{equation}
	\begin{split}
	( \textbf{F}_{S \rightarrow B}, \textbf{F}_{S \rightarrow R}, M ) = \mathcal{MI} \left(B, R \right),
	\end{split}
\end{equation}
where $ \textbf{F}_{S \rightarrow B} = \{ F_{S^t \rightarrow B}, t \in 0,\cdots,N-1 \} $ are the intermediate flows from targeted latent frames to the blurry input. Similarly, $\textbf{F}_{S \rightarrow R}$ denotes counterparts from latent frames to the RS view and $M$ is a predicted mask to aggregate warped frames using latent motion flows. The frame reconstruction part is implemented through \emph{GenNet}  in an encoder-decoder architecture. We propose a novel self-prompter based on motion residues to precisely and adaptively refine the warped latent frames,  which is formulated as:
\begin{equation}
	\begin{split}
	\textbf{S} = GenNet \left(B, R, \textbf{F}_{S \rightarrow B}, \textbf{F}_{S \rightarrow R}, M \right).
	\end{split}
	\vspace{-1mm}
\end{equation}
where $\textbf{S} = \{S^t, t \in 0,\cdots,N-1 \} $ denotes recovered latent sharp video sequence with a length of $N$.

\subsubsection{Dual Streams with Mutual Incentive}

Previous approaches that exploit temporal cues from adjacent frames~\cite{zhong2022animation,zhong2022blur} or dual reversed views from different cameras~\cite{zhong2022bringing,shang2023self} typically rely on straightforward feature concatenation to extract motion information. However, such strategies often fail to fully exploit the complementary characteristics of inputs with different temporal and exposure properties. In our cross-shutter setting, RS and blur views play inherently contrasting roles in reenacting the imaging moment, which needs a more nuanced treatment.

To this end, we propose a bifurcated processing scheme that explicitly disentangles the problem into parallel branches. As shown in \figref{model}b, the $i^{th}$ Motion Interpretation Block (MIB$_i$) adopts a dual-stream architecture with mutual incentives, facilitated by a shutter alignment and aggregation module. The RS branch focuses on capturing fine-grained local details and resolving motion direction ambiguities, while the blur branch leverages global temporal integration to provide robust context for motion magnitude estimation and to alleviate the ill-posedness of RS inputs (e.g., initial-state ambiguity). By capitalizing on the complementary strengths of the two exposure patterns, this design enables more accurate and robust motion field estimation.

To encourage meaningful interaction between two branches, we introduce a feature alignment and aggregation mechanism. Rather than concatenating the encoded features, we first predict bidirectional displacement fields between the two input views, denoted as $F^{i}_{B \rightarrow R}$ and $F^{i}_{R \rightarrow B}$. These flow maps are then used to warp the respective features, yielding aligned feature representations $\phi^{i}_{alin\_R}$ and $\phi^{i}_{alin\_B}$:
\begin{equation}
	\begin{split}
		&\phi^{i}_{alin\_R} = \mathcal{W} \left(\phi^{i}_B, F^{i}_{R \rightarrow B} \right)\\
		&\phi^{i}_{alin\_B} = \mathcal{W} \left(\phi^{i}_R, F^{i}_{B \rightarrow R} \right),
	\end{split}
\end{equation}
where $\mathcal{W}$ denotes backward-warping process. $\phi^{i}_B$ and $\phi^{i}_R$ are represented feature of blur and RS views. 
This alignment process allows the model to adaptively retain informative features while suppressing misleading or inconsistent signals from the complementary view. Additionally, the aggregated feature $\phi^{i}_{agg}$ is incorporated as an auxiliary input to further enhance motion interpretation. 

Therefore, taking the blur branch as an example, which can be formulated as:
\begin{equation}
	\begin{split}
		&\phi^{i}_B = \Phi_B \left(B, \textbf{F}^{i-1}_{S \rightarrow B} \right) \\
		&\textbf{F}^{i}_{S \rightarrow B} = \Psi_B \left( \left[ \phi^{i}_B, \phi^{i}_{alin\_B}, \phi^{i}_{agg} \right] \right),
	\end{split}
\end{equation}
where $\Phi_B $, $\Psi_B$ are encoder and decoder. $\left[ \cdot\right]$ denotes concatenation. The processing is similar under RS branch except for additional input, temporal positional encoding $\textbf{E}$. Overall, the $i^{th}$ MIB can be described as:
\begin{equation}
    \begin{split}
        \textbf{F}^{i},  F^{i}, M^i = \mathcal{MIB}_i \left(B,R,\textbf{E},\textbf{F}^{i-1},F^{i-1}, M^{i-1} \right),
    \end{split}
\end{equation}
where intermediate flows $\textbf{F}^i = [\textbf{F}_{S \rightarrow B}^i, \textbf{F}_{S \rightarrow R}^i]$, bidirectional flows $ F^i = [F_{B \rightarrow R}^i, F_{R \rightarrow B}^i] $ and $M^i$ is the predicted mask to aggregate warped frames. Following the strategy in RIFE~\cite{huang2022real}, MIB$_1$ predicts an initial flow directly from the image features without using any flow input. Iterative refinement starts from MIB$_2$, where each subsequent block refines the flow passed from the previous stage. This design avoids introducing external flow priors or poorly estimated initial flows that could propagate errors across iterations.

By aligning and aggregating the features across both branches, the network is able to cross-reference spatial and temporal evidence, further produce a more accurate and stable motion representation.

\subsubsection{Temporal Positional Encoding}
To further strengthen the model's capacity to tackle the motion ambiguity in latent video frames, we propose a temporal positional encoding specifically tailored for the RS branch within the MIB framework. The row-by-row exposure pattern of RS cameras intrinsically encodes temporal dynamics into the spatial domain of captured observations, as each row is sampled at a distinct time instant. The temporal offset across rows effectively imprints motion information within the RS image itself.

Inspired by the approximation of physical process for RS effects~\cite{liu2020deep}, we encode such temporal dynamics explicitly by assigning positional encodings to both the RS input $R$ and each latent frame $S^t$ defined as:
\begin{equation}
	\begin{split}
		[E_R]_k &= k, k=0,1,\cdots,N-1 \\
		E_{S^t} &= \frac{H-1}{N-1}t \cdot \mathbbb{1}.
	\end{split}
	\vspace{-1mm}
\end{equation}
Here, $[\cdot]_k$ denotes the operation that extracts $k^{th}$ row, $\mathbbb{1}$ is a 2-D tensor filled with ones and having the same spatial dimensions as the images, $H$ represents the image height, and $N$ is the number of recovered latent frames. While $E_R$ captures the row-wise readout time for each pixel in the RS image, $E_{S^t}$ represents the uniform temporal index of the latent frame, scaled to match the row index range.

Rather than directly using these absolute positional encodings, we further compute a relative temporal encoding between the RS image and each latent frame to better model the time offset between captured and latent content:
\begin{equation}
	\begin{split}
     \textbf{E} =\{ (E_R - E_{S^t}), t = 0,1,\cdots, N-1 \}.
	\end{split}
	\vspace{-1mm}
\end{equation}
This relative encoding implies the temporal distance between the observed RS image and each candidate latent frame, facilitating more accurate temporal reasoning in the reconstruction process. Finally, the resulting positional encoding maps $\textbf{E}$ are concatenated to the RS input $R$ along the channel dimension and fed into the RS branch of the MIB module, thereby enriching the input with explicit temporal structure to guide motion disambiguation.

\subsubsection{Shutter Alignment and Aggregation}
The Shutter Alignment and Aggregation (SAA) module facilitates comprehensive information exchange between two feature streams derived from distinct input perspectives, leveraging both aggregated and aligned feature representations. Specifically, the aggregated feature, denoted as $\phi_{agg}$, is computed by first concatenating the bidirectional displacement maps and then processing this concatenated input through a series of Encoder Blocks (EB). These encoder blocks are designed to effectively model inter-view correlations, allowing the network to capture shared structural and motion-related information across both views. In parallel, the aligned features $\phi_{alin\_B}$ and $\phi_{alin\_R}$ selectively incorporate reciprocal information from the opposite stream, enhancing contextual representation and temporal abstraction by emphasizing complementary cues across views.

The SAA module primarily comprises two separate encoders, each implemented using convolutional layers following a multi-output strategy, inspired by the design of MIMO-UNet~\cite{cho2021rethinking} (\figref{model}c). To account for spatial misalignment and visual distortions between the input views, two Spatial Transformer Networks (STNs) are employed as auxiliary branches within the SAA module. The STN branch is designed to capture coarse global transformation cues that complement the flow-prediction branch. Specifically, each STN predicts a global transformation from the encoder features, which is applied to spatially transform the corresponding feature maps. These globally aligned features are then aggregated with features from the flow branch, providing complementary global context for subsequent pixel-level displacement estimation. A connection is built through downsampling coarse outputs to the next block and the aggregated feature $\phi_{agg}$ is extracted from the output of the second encoder. The inference process within the $i^{th}$ SAA module can be formally expressed as:
\begin{equation}
	\begin{split}
		 F^i, M^i= \mathcal{SAA}_i \left(B,R, F^{i-1},M^{i-1} \right),
	\end{split}
	\vspace{-1mm}
\end{equation}
where $B$ and $R$ represent two input frames, and $F^{i-1}, M^{i-1}$ denote the bidirectional motion fields and occlusion mask from the previous stage.

\begin{figure*}[!tb]
	\centering
	\mfigure{1}{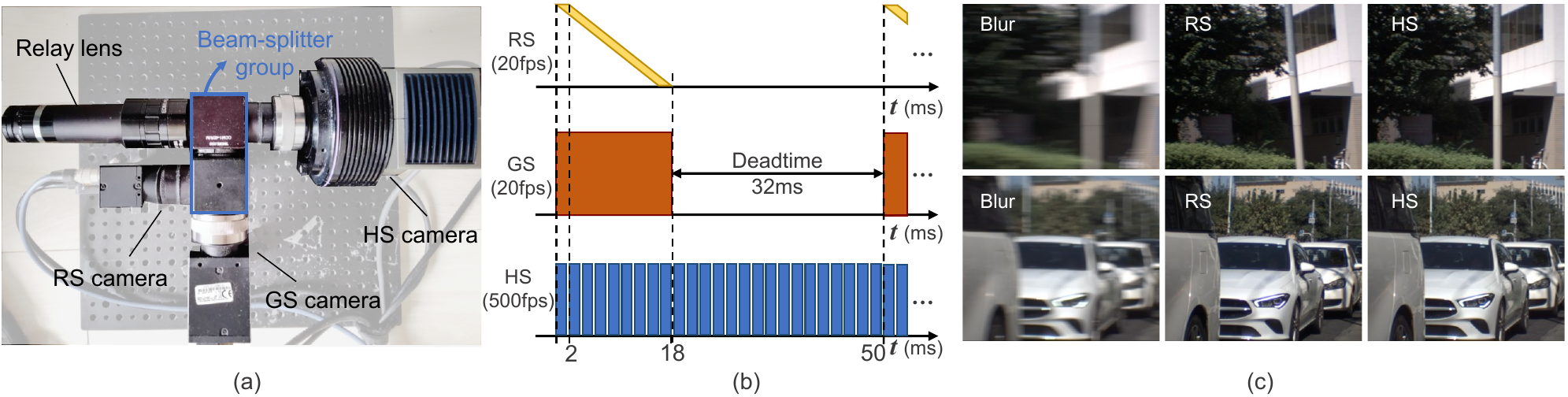}
	\vspace{-6mm}
	\caption{
		\textbf{Our triaxial imaging system.} (a) A photo of the actual system for data gathering; (b) Illustration of exposure duration for all cameras on temporal axis. In picture (b), its vertical axes can be interpreted as spatial rows of captured images from each camera; (c) shows representative samples from our real-world collection.
	}
	\label{fig:imaging_system}
	\vspace{-3mm}
\end{figure*}

\subsubsection{Motion-residue Prompted Frame Reconstruction}

In the warping–refinement paradigm, most methods adopt U-shaped architectures to capture multi-scale context, yet treat all regions uniformly~\cite{oh2022demfi,zhong2022animation,fan2022context}. This neglects the spatially varying reliability of warped results and can propagate errors under complex motion or occlusion. On the other hand, recent work explores prompt-based guidance for spatially varying degradations. For example, Wang \etal~\cite{wang2024selfpromer} propose a self-prompt dehazing Transformer based on depth difference maps. However, it relies on external depth estimation and ground-truth-dependent cues, increasing complexity and limiting applicability. These limitations motivate a more efficient and self-contained prompting strategy for motion-aware reconstruction.

Therefore, we propose a novel motion-residue prompted frame reconstruction module.  Instead of resorting to depth information, we build our self-prompter based on the difference between intermediate flows towards Blur and RS views, which highlights regions with complex motion and occlusion, exactly where the geometry structure and local textures need to be further corrected and refined (as shown in~\figref{prompt_visualization}). Compared with ~\cite{wang2024selfpromer}, our approach offers the following advantages: (1) it eliminates the need for additional depth estimation, thereby reducing computational overhead; and (2) it avoids iterative optimization during the testing phase by leveraging motion residues computed from intermediate flows. To inject the prompt, the flow difference maps $\textbf{F}_{diff} = \textbf{F}_{S \rightarrow B}-\textbf{F}_{S\rightarrow R}$ are first encoded by a Prompt Generation Network (PGN) to produce multi-scale outputs. These outputs are then integrated with the corresponding warped features by the Context Generation Network (CGN). The entire process is formulated as follows:
\begin{equation}
	\begin{split}
		&F_{cxt}^j= CGN\left(B,R,\textbf{F}_{S \rightarrow B},\textbf{F}_{S\rightarrow R} \right),\\
		&F_{prmpt}^j= F_{cxt}^j + F_{cxt}^j \times PGN\left( \textbf{F}_{diff} \right),
	\end{split}
\end{equation}
where $j=[0,1,2,3]$ is the scale level index and $F_{cxt}, F_{prmpt}$ denote extracted context feature and final prompt embedding. PGN is implemented following the structure of CGN in ~\cite{huang2022real}.

\subsection{Learning Objectives}
To further improve motion estimation accuracy and mitigate potential error accumulation during iterative refinement, we introduce a privileged distillation scheme. Recent VFI methods~\cite{huang2022real,kong2022ifrnet,li2023multi} have shown that optical flow distillation can significantly enhance performance. Kong~\etal~\cite{kong2022ifrnet} introduce a distillation loss function that selectively transfers beneficial knowledge from pre-trained optical flow models. However, these off-the-shelf flow priors are often misaligned with restoration tasks, leading to sub-optimal representations. Huang~\etal~\cite{huang2022real} further explore privileged distillation by allowing a teacher module to access ground-truth intermediate frames to supervise optical flow learning. Nevertheless, insufficient regularization causes the teacher to overuse privileged information, which undermines effective knowledge distillation~\cite{li2023multi}. To address these issues, we propose a region-adaptive motion distillation framework that mitigates overfitting by preventing the teacher model from excessively relying on privileged knowledge. Furthermore, instead of adopting a generic pre-trained optical flow model, we design a task-specific motion interpretation module as the teacher (MIB$_T$), which learns optical flow tailored to the proposed task.

Following the structure of student MIBs, we construct our teacher MIB$_T$ that is privileged to take the ground-truth latent frames and the outputs of MIB$_3$ as inputs:
\begin{equation}
	\begin{split}
		\textbf{F}^{T},  F^{T}, M^T = \mathcal{MIB}_T \left(B,R,\textbf{G},\textbf{E},\textbf{F}^{3},F^{3}, M^{3} \right),
	\end{split}
\end{equation}
where intermediate flows $\textbf{F}^T = [\textbf{F}_{S \rightarrow B}^T, \textbf{F}_{S \rightarrow R}^T]$, bidirectional flows $ F^T = [F_{B \rightarrow R}^T, F_{R \rightarrow B}^T] $ and ground-truth latent frames $\textbf{G} = \{G^t, t \in 0,\cdots,N-1 \} $.

To regularize the distillation process and make student modules focus on critical components, we introduce three types of masks, $\textbf{M}_d$, $\textbf{M}_b$ and $\textbf{M}_e$, actively addressing the distillation on highly dynamic areas, object boundaries and low-confidence estimate:
\begin{align}
	\begin{split}
		\textbf{M}_d = \mathbb{I} \left[  \|\mathbf{F}^T\|_2 > Q_3 + k \cdot (Q_3 - Q_1)  \right],
	\end{split} \\
	\begin{split}
		\textbf{M}_b = \mathrm{Norm} \left[ \sqrt{ \left(S_x \ast \textbf{G} \right)^2 + \left(S_y \ast \textbf{G} \right)^2 } \right],
	\end{split} \\
	\begin{split}		
		\textbf{M}_e = \mathbb{I} \left[ \| \tilde{\textbf{S}} - \textbf{G} \|_1 > \| \tilde{\textbf{S}}_T - \textbf{G} \|_1 \right],
	\end{split} \\
	\begin{split}		
		\textbf{M} = \mathrm{mean} \left[ \textbf{M}_d, \textbf{M}_b, \textbf{M}_e \right],
	\end{split}
\end{align}

where $\mathbb{I}[\cdot]$ denotes the indicator function, which returns 1 if the specified condition holds true, and 0 otherwise. The term $\|\mathbf{F}^T\|_2$ represents the Euclidean norm (i.e., magnitude) of the intermediate flows, with $Q_1$ and $Q_3$ denoting its first and third quartiles. The empirical coefficient $k$ is set to 2. $S_x$ and $S_y$ refer to the Sobel kernels applied along the x- and y-directions. $\mathrm{Norm}[\cdot]$ denote the min-max normalization applied to gradient–based quantity, ensuring all masks are expressed on a consistent scale. $\tilde{\mathbf{S}}$ and $\tilde{\mathbf{S}}_T$ are the warped latent frames, obtained using the flow fields $\mathbf{F}^3$ and $\mathbf{F}^T$, respectively. We finally determine the weights for three masks via grid search on the validation set.

Therefore, the distillation loss is formulated as:
\begin{equation}
	\begin{split}
		\mathcal{L}_{dis} = \textbf{M} \cdot \| \textbf{F}^3 - \textbf{F}^T\|_1,
	\end{split}
\end{equation}
combined with the reconstruction loss:
\begin{equation}
	\begin{split}
		&\mathcal{L}_{rec} = \sqrt{\| \textbf{S} -\textbf{G} \|^2 + \epsilon^2}, \\
		&\mathcal{L}_{rec}^T = \sqrt{\| \tilde{\textbf{S}}_T -\textbf{G} \|^2 + \epsilon^2},
	\end{split}
\end{equation}
where $\epsilon$ is a constant which we empirically set to $10^{-3}$ for all the experiments. Finally, we train our model with:
\begin{equation}
	\begin{split}
		\mathcal{L}_{total} = \mathcal{L}_{rec} + \mathcal{L}_{rec}^{T} + \lambda_d \mathcal{L}_{dis}.
	\end{split}
\end{equation}
$\lambda_d$ is a hyper-parameter to balance the distillation and reconstruction loss.

\section{Optical System and Datasets}
\label{sec:data}


\subsection{A Triaxial Imaging System } 
\label{sec:imaging_system}

\begin{table*}[!tp]
	\centering
	\caption{\textbf{The comparisons of our realBR dataset compared with other existing dataset.} We denote supported tasks as: (a) RS correction, (b) deblurring, (c) blur decomposition, and (d) RS Temporal Super-resolution. }
	\label{tab:dataset_cmp}
	\vspace{-2mm}
	\resizebox{0.95\linewidth}{!}
	{
		\begin{tabular}{@{}lcccccccc@{}}
			\toprule
			& Carla~\cite{liu2020deep} & Fastec~\cite{liu2020deep} & Gev-RS~\cite{zhou2022evunroll} & BS-RSC~\cite{cao2022learning} & GOPRO~\cite{nah2017deep} & RBI~\cite{zhong2022blur}  & Adobe240~\cite{shen2020video} & realBR\\
			\midrule
			Type       & synthetic     & synthetic      & synthetic           & real            & synthetic       & real           & synthetic & real \\
			Mode       & RS            & RS             & Event, RS      & RS              & Blur            & Blur           & Blur   & RS, Blur \\
			
			Scenes     & 250           & 76             & 29             & 80              & 33              & 55             & 120  &  84  \\
			Frames     & 2,500         & 2,584          & 3,700          & 4,000           & 3,214           & 1,375          & 14,049 & 9,084 \\
			GTs        & 5,000         & 5,168          & 3,700          & 4,000           & 32,140    & 27,500         & 114,747   & 93,600 \\
			Tasks       & a, d          & a, d           & a              & a               & b, c            & b, c           & b, c & a, b, c, d \\
			Resolution & 640$\times$480& 640$\times$480 & 640$\times$360 & 1024$\times$768 & 1280$\times$720 & 640$\times$480 & 640$\times$352 & 800$\times$800  \\
			\bottomrule
		\end{tabular}
	}
	\vspace{-3mm}
\end{table*}

\begin{table}[!tp]
	\centering
	\caption{\textbf{Qualitative alignment assessment of our imaging system.} We compute pixel-level difference and accuracy between RS and Blur inputs with respect to GT.}
	\label{tab:dataset_evaluation_numerical}
	\resizebox{1\linewidth}{!}
	{
		\begin{tabular}{rccccc}
			\toprule
			\multirow{2}{*}{Input}& \multicolumn{2}{c}{Error} & \multicolumn{3}{c}{Accuracy} \\ \cmidrule(lr){2-3}\cmidrule(lr){4-6}
			
			&  MSE     & Abs.rel    & $\delta < 1.15$    & $\delta < 1.25$ & $\delta < 1.35$  \\ \midrule
			GS view &    0.0003 & 0.0448 & 0.9587 & 0.9878 &  0.9949    \\
			
			RS view &    0.0003   & 0.0410  & 0.9813 & 0.9962 & 0.9993    \\ \bottomrule
		\end{tabular}
	}
	\vspace{-4mm}
\end{table}


Existing blur and rolling-shutter benchmarks are largely synthetic, relying on frame averaging~\cite{shen2020blurry,oh2022demfi,shen2020video} or scanline sampling~\cite{liu2020deep,zhou2022evunroll} from high-speed datasets such as GOPRO~\cite{nah2017deep} and Adobe240~\cite{shen2020video}. These approaches often fail to capture real-world degradations, introducing unnatural artifacts~\cite{rim2020real,cao2022learning}. Moreover, no rigorous studies explore the potential of shutter combination to simultaneously handle blur and RS effects.

Inspired by recent optical system design~\cite{rim2020real,zhong2022blur}, we develop a triaxial imaging setup (Figure~\ref{fig:imaging_system}a) capable of capturing synchronized training inputs (RS effect, GS blur), and high-speed ground truth sequences. The setup utilizes two beam-splitters to divide incoming light into three identical optical paths, each directed toward a dedicated camera: an RS camera (FLIR BFS-U3-63S4C with 2×2 binning), a GS camera (FLIR GS3-U3-23S6C), and a high-speed (HS) camera (BITRAN CS-700C with forced cooling). 
To achieve precise alignment, we first mount the RS camera as a reference and iteratively adjust the GS and HS cameras using residual checkerboard images. The system is then calibrated using two homographies under a closed-loop constraint~\cite{rim2020real}. We further validate this strategy on 11 static scenes without camera or object motion.
~\tabref{dataset_evaluation_numerical} reports MSE, Abs Rel and accuracy. Abs Rel $=\frac{1}{|P|}\sum_{p\in P}\frac{|d_p-d^{gt}_p|}{d^{gt}_p}$, and
accuracy is the percentage of pixels with
$\delta=\max(d_p/d^{gt}_p,\, d^{gt}_p/d_p)<\text{thr}$,
where $d_p$ and $d^{gt}_p$ denote the input and GT values
at pixel $p$, and $P$ is the set of valid pixels. The results validate that our system can capture well-aligned data with high precision.

Additionally, to ensure consistent light integration, a neutral density (ND) filter with 20\% transmittance is applied in the GS branch to compensate for its longer exposure time, while a second ND filter harmonizes illumination between the RS and blur views. This balances exposure time and incoming light intensity across sensors, ensuring photometric consistency, similar to~\cite{duan2025eventaid}. Detailed hardware specifications and optical configurations are provided in the supplementary material.

\begin{table*}[!tp]
	\centering
	\caption{\textbf{Quantitative comparison with SOTAs on realBR.} We evaluate the performance of reconstructing latent frame sequence with lengths of $3$, $5$ and $9$. `\emph{B}$\cdot$\emph{R}' is our proposed dual blur-RS view, `\emph{n}$\cdot$\emph{B}' is the setting using \emph{n} neighboring blur frames to tackle motion ambiguity and `\emph{n}$\cdot$\emph{R}' is instead using \emph{n} neighboring RS frames to tackle initial-state ambiguity. 
	}
	\label{tab:compare_with_sota}
	\resizebox{0.97\linewidth}{!}
	{
		\begin{tabular}{rccccccccccccc}
			\toprule
			\multirow{2}{*}{Method}& \multirow{2}{*}{Input} & \multicolumn{3}{c}{$\times$3} & \multicolumn{3}{c}{$\times$5} & \multicolumn{3}{c}{$\times$9} & \multirow{2}{*}{\begin{tabular}[c]{@{}l@{}}Time  \\ \,\,\,\,(s)\end{tabular}} & \multirow{2}{*}{\begin{tabular}[c]{@{}l@{}}Params\\    \quad(M)\end{tabular}} & \multirow{2}{*}{\begin{tabular}[c]{@{}l@{}}FLOPs\\ \quad(G)\end{tabular}} \\ \cmidrule(lr){3-5}\cmidrule(lr){6-8}\cmidrule(lr){9-11}
			
			& & PSNR     & SSIM     & LPIPS    & PSNR    & SSIM    & LPIPS   & PSNR    & SSIM    & LPIPS   &      &        &   \\ \cmidrule(lr){1-14}
			
			LEVS~\cite{jin2018learning}& \emph{1}$\cdot$\emph{B} & 21.77 & 0.7042 & 0.2886 & 21.62 &  0.7153 & 0.2683 & 21.83 & 0.7277 & 0.2535 &  1.47 & 15.9 & 304  \\ \cmidrule(lr){1-14}
			
			AfB$_p$ ~\cite{zhong2022animation}&  \multirow{4}{*}{\emph{2}$\cdot$\emph{B}}  & 21.50 &  0.7596 & 0.4102 & 21.65 & 0.7648  & 0.4055 & 21.82  & 0.7686 & 0.4017 & 0.15  &  190 & 839  \\
			
			AfB$_v$ ~\cite{zhong2022animation}&   & 22.83 &  0.7877 & 0.3904 & 22.96 & 0.7903  & 0.3883 & 23.10  & 0.7924 & 0.3860 & 0.22  & 129 & 793  \\
			
			RIFE$_{B}$~\cite{huang2022real} &   & 24.60  & 0.8172 & 0.2254  & 24.73 & 0.8199 & 0.2268 & 24.83 & 0.8219 & 0.2268 &  1.33   & 54.8 & 71.1    \\
			
			IFED$_{B}$~\cite{zhong2022bringing} &  & 24.45  & 0.8105 & 0.1817  & 24.62 & 0.8141 & 0.1811 & 24.74 & 0.8164 & 0.1798 & 1.33 & \cellcolor{second}10.8  & \cellcolor{best}29.5    \\ \cmidrule(lr){1-14}
			
			BiT ~\cite{zhong2022blur}&  \emph{3}$\cdot$\emph{B}  & 21.90 & 0.7664  & 0.2583 & 21.88 & 0.7694 &  0.2574 & 22.02  & 0.7729 & 0.2546 & \cellcolor{best}0.11 &  11.3 & \cellcolor{second}57.4 \\ \cmidrule(lr){1-14}

			DeMFI~\cite{oh2022demfi} &  \emph{4}$\cdot$\emph{B} & 25.55  & 0.8485 & 0.2247  & 25.26 & 0.8466 & 0.2275 & 26.20  & 0.8577 & 0.2165 &  4.86   &  \cellcolor{best}7.41  & 420    \\  \cmidrule(lr){1-14}
			
			
			RSSR ~\cite{fan2021inverting}&  \multirow{5}{*}{\emph{2}$\cdot$\emph{R}}  & 22.25 & 0.7702  & 0.1507 & 21.92 &  0.7633 & 0.1526 & 22.82 & 0.7833  & 0.1362     & 2.19 & 26.0 &42.7 \\
			
			CVR ~\cite{fan2022context}&   &21.98  &  0.7690 & 0.1618 & 21.70 &  0.7620 & 0.1717 & 22.26 & 0.7761  & 0.1564 & 2.56  & 42.7 & 101  \\
			
			RIFE$_{R}$~\cite{huang2022real} &  &  23.70 &0.8009  & 0.1998  & 24.14  & 0.8124 & 0.1875  & 24.36 & 0.8181 & 0.1813 & 1.33   & 54.8 & 71.1  \\
			
			IFED$_{R}$~\cite{zhong2022bringing} &   & 23.93  &0.7935  & 0.1114  & 24.33  & 0.8033 & 0.1032  & 24.54 & 0.8085 & 0.0989 & 1.33 & 10.8  & 29.5 \\ 
			
			LBCNet~\cite{fan2024learning} &  & 21.17  & 0.7582 & 0.1685  &  21.20  & 0.7595 & 0.1689  & 21.19 & 0.7593 & 0.1705 &   0.23  &  10.7  &  53.6  \\ \cmidrule(lr){1-14}
			
			RIFE$_{BR}$~\cite{huang2022real} &  \multirow{4}{*}{\emph{B}$\cdot$\emph{R}}   & 30.26  & 0.8983 & 0.1071  & 30.53 & 0.9030 & 0.1046 & 30.67 & 0.9053 & 0.1042 &  1.33   & 54.8   & 71.1    \\
			
			IFED$_{BR}$~\cite{zhong2022bringing} &   & 30.46  & 0.9030 & 0.0467  & 30.70 & 0.9064 & 0.0445 & 30.84 & 0.9084 & 0.0434 & 1.33  & 10.8   & 29.5    \\
			Ours-lite \,\,\, &  &  30.70 & 0.9023 & 0.0670  & 30.89  &0.9058  &  0.0657 & 31.00 & 0.9075  & 0.0650  & 0.49 & 59.4 & 96.8  \\
			Ours \quad\,\,\,\, &  &  30.97 & 0.9073 & 0.0695  & 31.19  &0.9105  &  0.0675 & 31.31 & 0.9122 & 0.0671 &  1.31  & 120   & 187  \\\bottomrule
		\end{tabular}
	}
\end{table*}

\begin{figure*}[!tb]
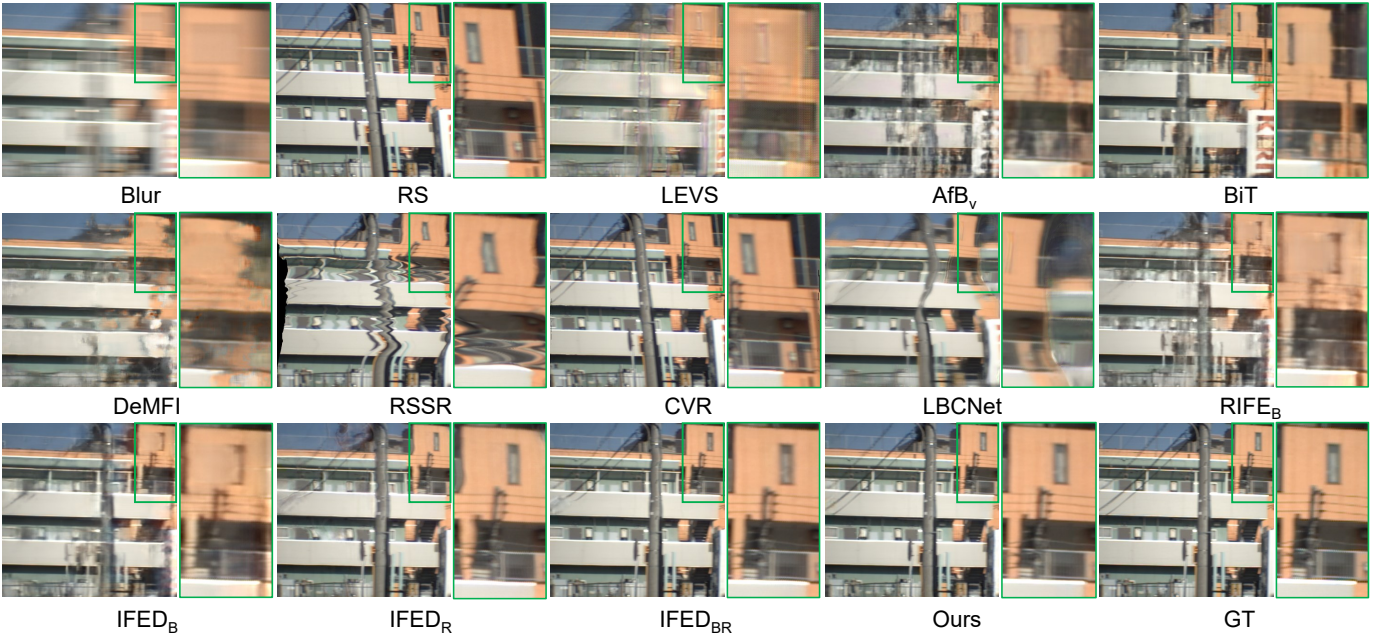

	\vspace{-2mm}
	\centering
	\mfigure{1}{exp/compare_with_sota_v4.pdf}
	\vspace{-4mm}
	\caption{
		\textbf{Qualitative comparison with SOTAs on realBR.} Our model outperforms the approaches approximating latent motion fields relying on adjacent blurry or RS inputs.
	}
	\label{fig:compare_with_sota}
	\vspace{-6mm}
\end{figure*}

\subsection{The Captured RealBR Dataset} 
\label{sec:data_collect}

We apply our optical system to construct the realBR (GS Blur \& RS) dataset, which comprises 84 street scenes with diverse objects and motions. The dataset is acquired under two capture settings. The first setting includes 54 scenes, from which we obtain 56 consecutive RS–GS blur pairs and 1,400 aligned high-speed (HS) sharp frames. As illustrated in Fig.~\ref{fig:imaging_system}b, each RS–GS pair is captured within a $50\,\mathrm{ms}$ period, during which both GS and RS cameras complete their $18\,\mathrm{ms}$ exposure, followed by a $32\,\mathrm{ms}$ deadtime. Meanwhile, the HS camera operates at $500\,\mathrm{fps}$, capturing 25 frames with a $2\,\mathrm{ms}$ exposure, among which 9 fall within the GS/RS exposure window. After preprocessing, this portion of data is split into 40 training, 4 validation, and 10 test scenes. Representative samples are shown in Fig.~\ref{fig:imaging_system}c. The remaining 30 scenes are captured under a different setting for third-party validation. Each scene contains 60 RS–GS blur pairs and 600 corresponding HS sharp frames. As summarized in Table~\ref{tab:dataset_cmp}, realBR is the first real-world dataset providing spatially and temporally aligned RS effects, GS blur, and high-speed sharp frames, enabling comprehensive benchmarking for motion degradation inverting tasks.

\subsection{The Synthesized GOPRO-BR Dataset} 
\label{sec:data_synthetic}
To further validate our findings on reverting motion degradation using the complementary Blur and RS observations, we also construct a synthetic dataset, GOPRO-BR. This dataset is generated following the methodology in~\cite{cho2021rethinking,oh2022demfi,tao2018scale}, based on the GOPRO dataset~\cite{nah2017deep}, which contains 33 video sequences at a resolution of $1280 \times 720$, each comprising 1200 frames captured at $240\, \mathrm{fps}$. To simulate more realistic motion effects, we first apply a $\times64$ interpolation to the original GOPRO frames using a state-of-the-art VFI method~\cite{huang2022real}. The interpolated sequences are then center-cropped to a size of $512\times512$. Throughout the synthesis process, we strictly enforce the conditions specified in \eqnref{blur} and \eqnref{rs} to ensure that the resulting blur and RS frames are aligned and depict the same scene content.

	\section{Experiments}
\label{sec:experiments}

\begin{figure*}[!tb]
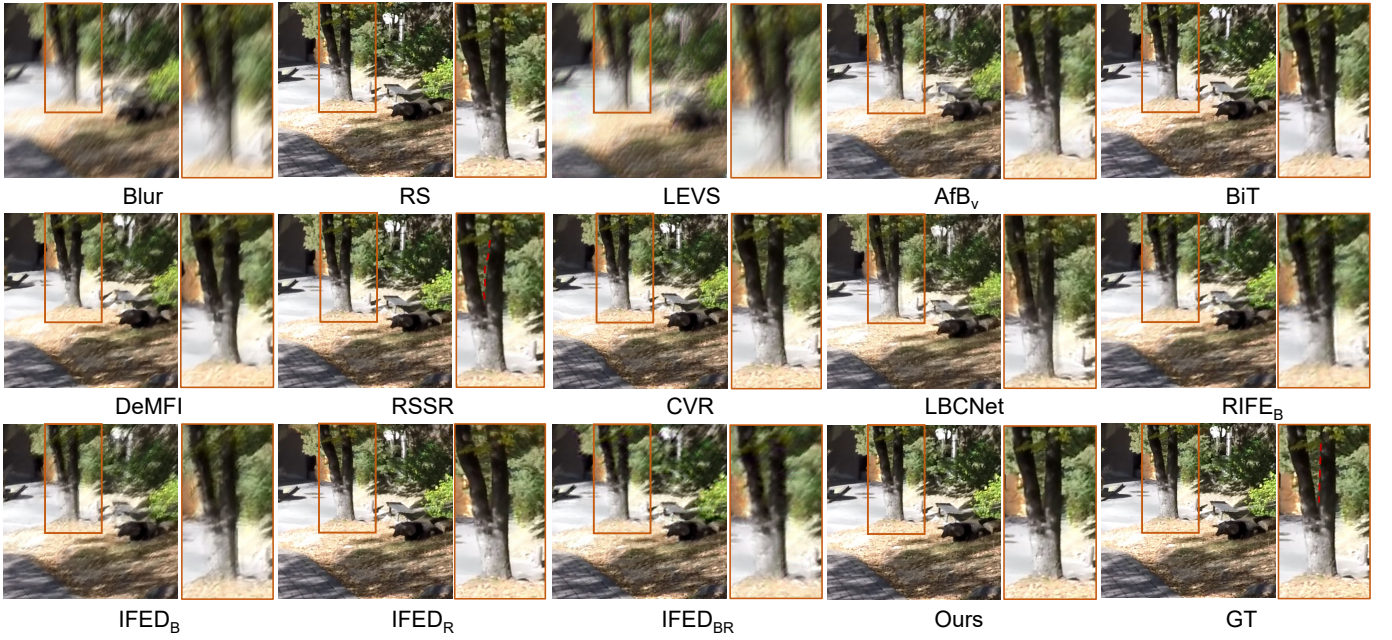

	\vspace{-1mm}
	\centering
	\mfigure{1}{exp/compare_with_sota_gopro_v1.pdf}
	\vspace{-5mm}
	\caption{
		\textbf{Qualitative comparison with SOTAs on GOPRO-BR.} Our model consistently outperforms all compared methods.
	}
	\label{fig:compare_with_sota_gopro}
	\vspace{-4mm}
\end{figure*}


We compare our model against existing SOTAs to handle motion ambiguity of blur decomposition, including LEVS~\cite{jin2018learning}, AfB~\cite{zhong2022animation}, and BiT~\cite{zhong2022blur}. For AfB, we consider two types of motion guidance: learned by a predictor(AfB$_p$), and extracted from adjacent blur frames (AfB$_v$). Although, we focus on recovering latent frames within exposure, we also incorporate DeMFI~\cite{oh2022demfi}, a blur frame interpolation method, by adapting it to our task for fairness. 
On the other hand, the leading approaches for RS temporal super-resolution, RSSR~\cite{fan2021inverting}, CVR~\cite{fan2022context} and LBCNet~\cite{fan2024learning} are also taken into consideration. To better demonstrate the superiority of our model and setup, we include methods with more competitive settings: IFED~\cite{zhong2022bringing} with dual reversed RS views; EvUnroll~\cite{zhou2022evunroll} and EBFI~\cite{weng2023event}, both assisted by event cameras; and PMB~\cite{rengarajan2020photosequencing}, which fuses short and long exposures. Because of lacking different modal data, we simulate event streams from high frame-rate latent videos by an event simulator~\cite{gehrig2020video}. For PMB, we strictly follow the experimental setup, adding noise to the first sharp GT frame to simulate a short-exposure observation and jointly using it with the long-exposure blurred view as input. Moreover, the synthesizing process of reversed RS frames is strictly aligned to original RS views~\cite{zhong2022bringing}. In addition, considering that RIFE~\cite{huang2022real}, where we choose its full-capacity version for a fair comparison, and IFED can be easily adapted to our blur-RS input, we therefore combine the two models with our proposed setting (denoted as RIFE$_{BR}$ and IFED$_{BR}$). As a contrast, results of these two models using consecutive blur frames (RIFE$_B$ and IFED$_B$) or RS frames (RIFE$_R$ and IFED$_R$) are also provided. Importantly, to ensure a fair evaluation across all baselines, we retrain all models using our collected datasets. 

\subsection{Comparison with SOTA methods}
\label{sec:comp_sota}

\begin{table}[!tp]
	\centering
	\caption{\textbf{Quantitative comparison with SOTAs on GOPRO-BR.} We evaluate performance for sequence lengths of $3$, $7$.}
	\label{tab:compare_with_sota_gopro}
	\resizebox{\linewidth}{!}
	{
		\begin{tabular}{rccccccc}
			\toprule
			\multirow{2}{*}{Method}& \multirow{2}{*}{Input} & \multicolumn{3}{c}{$\times$3} & \multicolumn{3}{c}{$\times$7} \\ \cmidrule(lr){3-5}\cmidrule(lr){6-8}
			
			& & PSNR     & SSIM     & LPIPS    & PSNR    & SSIM    & LPIPS    \\ \midrule
			
			LEVS~\cite{jin2018learning} & \emph{1}$\cdot$\emph{B} & 17.27 & 0.6063 & 0.3410 & 16.64 & 0.5800 & 0.3811  \\ \midrule
			
			AfB$_p$ ~\cite{zhong2022animation}& \multirow{4}{*}{\emph{2}$\cdot$\emph{B}}  & 23.38 &  0.7411 & 0.2271 & 23.41  & 0.7517 & 0.2183   \\
			AfB$_v$ ~\cite{zhong2022animation}&   & 28.10 &  0.8760 & 0.1496 & 28.39 & 0.8815 & 0.1461 \\
			
			RIFE$_{B}$~\cite{huang2022real} &    & 31.26 & 0.9410 & 0.0896  & 31.49 & 0.9430 & 0.0892    \\
			
			IFED$_{B}$~\cite{zhong2022bringing} &   & 29.46 &  0.9193 & 0.0897 & 29.75 & 0.9225 & 0.0874 \\ \midrule
			
			BiT ~\cite{zhong2022blur}& \emph{3}$\cdot$\emph{B}  & 32.31  & 0.9234 & 0.0708 & 32.56  & 0.9266 & 0.0691 \\ \midrule
			
			DeMFI~\cite{oh2022demfi} & \emph{4}$\cdot$\emph{B} & 27.57  & 0.9002 & 0.1332 & 27.44  & 0.8984 & 0.1304   \\  \midrule
			
			

			RSSR ~\cite{fan2021inverting}& \multirow{5}{*}{\emph{2}$\cdot$\emph{R}}  & 22.73 &  0.8116 & 0.1039 & 22.65 & 0.8090  & 0.1154   \\ 
			
			CVR ~\cite{fan2022context}&   & 23.50 &  0.8342 & 0.0818 & 23.47 & 0.8332  & 0.0815  \\ 
			
			RIFE$_{R}$~\cite{huang2022real} &   & 24.16 & 0.8318 & 0.1697   & 24.32 & 0.8365 & 0.1618 \\
			
			IFED$_{R}$~\cite{zhong2022bringing}& & 28.30 & 0.9122  & 0.0475 & 28.63 & 0.9181  &0.0446   \\      
			LBCNet~\cite{fan2024learning}& & 24.16 & 0.8434  & 0.0731 & 23.99 & 0.8384  &0.0758    \\ 
			\midrule
			
			

			RIFE$_{BR}$~\cite{huang2022real} &  \multirow{4}{*}{\emph{B}$\cdot$\emph{R}}  & 34.49 & 0.9701 & 0.0398   & 35.02 & \cellcolor{second}0.9733 & 0.0366    \\
			
			IFED$_{BR}$~\cite{zhong2022bringing} &   & 33.03  & 0.9627 & 0.0332 & 33.72 & 0.9675 &  0.0304    \\
			Ours-lite \,\,\,\, &  & 34.63 & 0.9714  & 0.0363  & 35.24 & 0.9749 & 0.0333   \\
			Ours\quad\,\,\,\,\, &  & 35.10 & 0.9743  & 0.0343  & 35.63 & 0.9770 & 0.0317    \\ \bottomrule
		\end{tabular}
	}
	\vspace{-3mm}
\end{table}

\heading{Experiments on RealBR}
\tabref{compare_with_sota} summarizes the quantitative comparisons with various methods. Overall, recovering a video from single blurred image remains highly challenging. The carefully crafted supervision strategy in LEVS\cite{jin2018learning} offers limited help in addressing the ambiguity caused by temporal averaging. In contrast, by predicting motion across adjacent blurred inputs, models are able to infer the temporal order of latent frames to a certain extent. Among such approaches, DeMFI~\cite{oh2022demfi} achieves the best performance but still falls significantly short of our model. On the other hand, methods like RSSR~\cite{fan2021inverting}, CVR~\cite{fan2022context}, and LBCNet~\cite{zhong2022bringing} utilize consecutive RS inputs to alleviate the initial-state ambiguity. However, the absence of sufficient global context leads to results that are even inferior to those obtained from blurred inputs. Remarkably, RIFE$_{BR}$~\cite{huang2022real} and IFED$_{BR}$~\cite{zhong2022bringing} achieve substantial improvements, exceeding 5.6 dB in PSNR, when compared to their single-modality counterparts (RIFE$_B$, IFED$_B$, RIFE$_R$, and IFED$_R$). These gains clearly demonstrate the effectiveness and complementarity of combining blur and RS views.
Furthermore, under the same dual RS-blur setting, our method outperforms both RIFE$_{BR}$ and IFED$_{BR}$, validating the superiority of our proposed model.

\begin{table}[!h]
	\centering
	\setlength{\tabcolsep}{4pt}
	\caption{\textbf{Numerical comparison of temporal consistency.} Lower tOF and tLP denote better coherence.}
	\vspace{-1mm}
	\resizebox{1\columnwidth}{!}{
		\begin{tabular}{l|cccccccc} 
			\toprule
			Methods & \makecell[c]{ AfB$_v$ \\ {\cite{zhong2022animation}}}  & \makecell{LBCNet \\ {\cite{fan2024learning}}}  & \makecell[c]{IFED$_R$ \\ {\cite{zhong2022bringing}}} & \makecell[c]{RIFE$_B$ \\ {\cite{huang2022real}}} & \makecell[c]{BiT \\ {\cite{zhong2022blur}}} & \makecell[c]{RIFE$_R$ \\ {\cite{huang2022real}}} & \makecell[c]{IFED$_{BR}$ \\ {\cite{zhong2022bringing}}} & Ours \\
			\midrule
			tOF $\downarrow$        & 2.42  & 1.51 & 0.86 & 0.52 & 0.45 & 0.58 & 0.43 & 0.40 \\
			tLP$\times100\downarrow$ & 4.41  & 2.46 & 1.55 & 1.63 & 1.39 & 1.64 & 0.68 & 0.68 \\
			\bottomrule
		\end{tabular}
	}
	\label{tab:temporal_numerical}
	\vspace{-2mm}
\end{table}

\begin{figure}[!th]
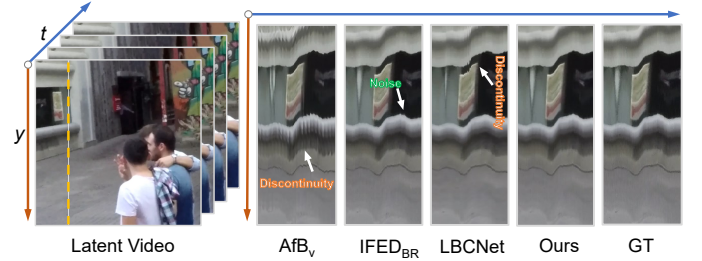

	\vspace{-2mm}
	\mfigure{1}{exp/temporal_profile_v2.pdf}
	\vspace{-6mm}
	\caption{
		\textbf{Comparison of temporal profiles.} We visualize the pixels of the
		selected columns (dotted line) according to ~\cite{chan2022basicvsr++}.
	}
	\label{fig:temporal_profile}
	\vspace{-5mm}
\end{figure}

\begin{table*}[!tp]
	\centering
	\caption{\textbf{Quantitative comparison with competitive settings on GOPRO-BR and realBR.} We evaluate the performance by metrics PSNR/SSIM/LPIPS. RS views are directly captured in realBR, it is unable to synthesize aligned inverse RS views. Therefore, we report IFED results only on the GOPRO-BR dataset.}
	\label{tab:compare_with_competitive_setting} 
	\resizebox{\linewidth}{!}
	{
		\begin{tabular}{rccccccc}
			\toprule
			\multirow{2}{*}{Method}& \multirow{2}{*}{Input} & \multicolumn{3}{c}{GOPRO-BR} & \multicolumn{3}{c}{realBR} \\ \cmidrule(lr){3-5}\cmidrule(lr){6-8}
			
			& & $\times$3     & $\times$5     & $\times$7    & $\times$3    & $\times$5    & $\times$9    \\ \midrule
			
			PMB~\cite{rengarajan2020photosequencing} & \multirow{3}{*}{\emph{B}$\cdot$\emph{SL}} & 35.48/0.9723/0.0349  &34.83/0.9699/0.0338  & 35.11/0.9715/0.0324 & 27.74/0.8560/0.1568 &  27.37/0.8510/0.1648 & 27.20/0.8489/0.1683 \\  
			RIFE$_{BS}$~\cite{huang2022real}  &   & 34.37/0.9660/0.0575 &34.42/0.9669/0.0576  & 34.32/0.9666/0.0586   & 29.67/0.8827/0.1431  & 29.61/0.8826/0.1459 &  29.59/0.8826/0.1474 \\
			IFED$_{BS}$~\cite{zhong2022bringing} &   & 32.92/0.9540/0.0501  & 33.09/0.9563/0.0475 &33.02/0.9562/0.0478  & 28.72/0.8602/0.1129  &  28.72/0.8614/0.1119 & 28.74/0.8623/0.1115     \\ \midrule
			
			EBFI~\cite{weng2023event}& \emph{B}$\cdot$\emph{Event} & 33.21/0.9568/0.0703 & 33.36/0.9581/0.0694  & 33.51/0.9591/0.0685 & 27.65/0.8698/0.1709 &  27.96/0.8751/0.1666 & 28.13/0.8781/0.1636   \\ \midrule
			
			IFED~\cite{zhong2022bringing}& \emph{R}$\cdot$\emph{iR} & 30.89/0.9417/0.0372 &31.40/0.9478/0.0337   &31.96/0.9530/0.0307  & -- &  -- & -- \\ \midrule
			
			EvUnroll~\cite{zhou2022evunroll}& \emph{R}$\cdot$\emph{Event} & 33.06/0.9558/0.0737 & 33.33/0.9575/0.0714  & 33.48/0.9587/0.0699  & 23.61/0.8145/0.2294 &  23.37/0.8093/0.2411 & 23.25/0.8069/0.2465 \\ \midrule

			RIFE$_{BR}$~\cite{huang2022real} &  \multirow{3}{*}{\emph{B}$\cdot$\emph{R}}  & 34.49/0.9701/0.0398 &34.84/0.9722/0.0378  &35.02/0.9733/0.0366    &30.26/0.8983/0.1071  & 30.53/0.9030/0.1046 &  30.67/0.9053/0.1042    \\
			
			IFED$_{BR}$~\cite{zhong2022bringing} &   & 33.03/0.9627/0.0332  & 33.43/0.9656/0.0316 &33.72/0.9675/ 0.0304  &30.46/0.9030/0.0467  &  30.70/0.9064/0.0445 & 30.84/0.9084/0.0434     \\
			
			Ours\quad\,\,\,\,\, &  & 35.10/0.9743/0.0343 & 35.44/0.9760/0.0326  & 35.63/0.9770/0.0317   & 30.97/0.9073/0.0695 & 31.19/0.9105/0.0675 & 31.31/0.9122/0.0671   \\ \bottomrule
		\end{tabular}
	}
	\vspace{-2mm}
\end{table*}

\begin{figure*}[!tp]
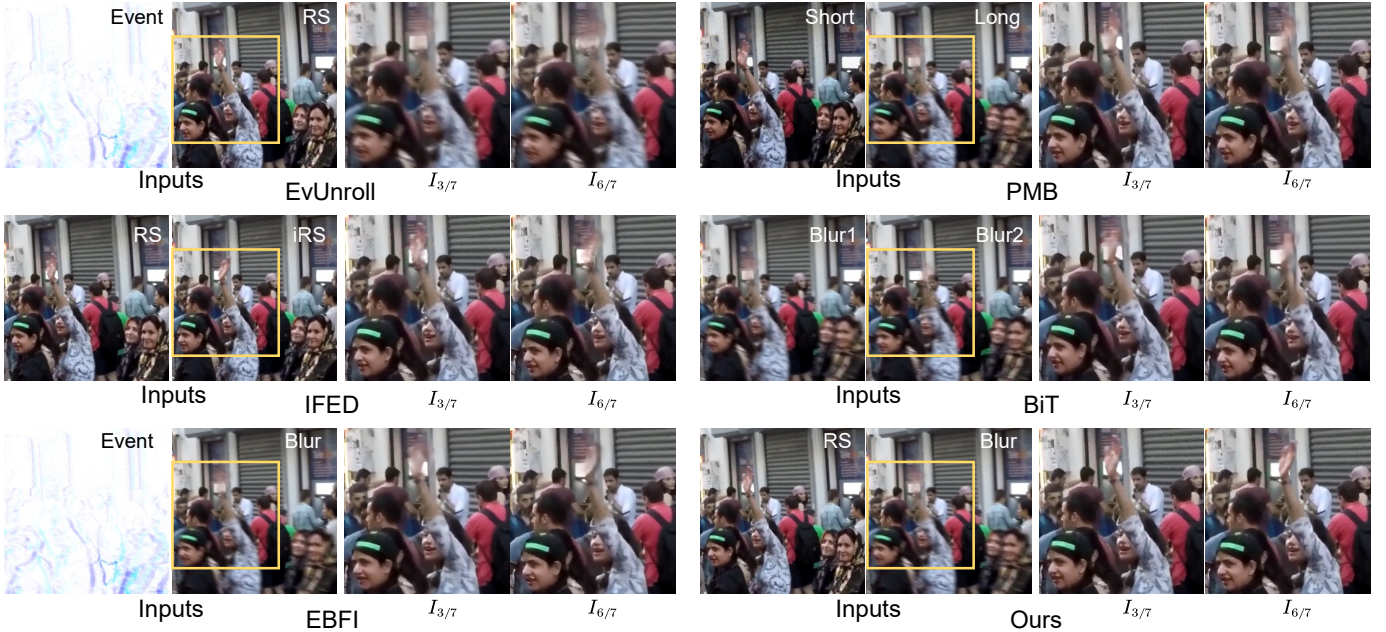

	\mfigure{1}{exp/compare_settings_gopro.pdf}
	\vspace{-6mm}
	\caption{
		\textbf{Qualitative comparison with competitive settings on GOPRO-BR.} $I_t$ denotes the recovered latent frame at time $t$.
	}
	\label{fig:compare_with_competitive_setting_gopro}
	\vspace{-4mm}
\end{figure*}

To better illustrate the experimental results, several representative restored frames are presented in~\figref{compare_with_sota}. While existing methods reduce distortions to some extent, they fail to fully recover fine local details and structures. In contrast, our results are markedly clearer and more closely aligned with GT, consistent with the quantitative findings. These observations further validate the superior effectiveness of our method in reconstructing latent videos from motion degradation, outperforming SOTA approaches.

\heading{Experiments on GOPRO-BR}
As a supplementary validation on realBR, we also conducted experiments on the synthetic dataset GOPRO-BR. Quantitative and qualitative results are presented in \tabref{compare_with_sota_gopro} and \figref{compare_with_sota_gopro}. Notably, although methods using consecutive RS inputs (e.g., RSSR, CVR, and IFED$_R$) yield visually appealing results in terms of recovered details, they fail to reconstruct the correct scene structure. As shown in the zoomed-in patches, the outline of the tree trunk is severely distorted, appearing curved rather than straight as in the GT, and the entire structure is noticeably shifted to the right. In contrast, our solution consistently outperforms all compared approaches in both correcting structural distortions and recovering fine details.

\heading{Temporal Consistency }
As shown in ~\figref{temporal_profile}, we compare the temporal profiles of our model with SOTA methods. The profile generated by IFED$_{BR}$ exhibits noticeable noise, suggesting the presence of flickering artifacts. Similarly, the profiles from LBCNet and AfB$_v$ still suffer from temporal discontinuities. In contrast, our method yields a smoother temporal transition by more effectively resolving the ambiguities inherent in motion degradation inversion. The quantitative results in~\tabref{temporal_numerical} are consistent with the qualitative observations and also demonstrate that our method achieves superior temporal consistency. Additional results and visual comparisons with a broader range of methods are available in \href{https://jixiang2016.github.io/dualBR_site/}{\textit{Video Demos}}, further highlighting the advantages of our approach, particularly in achieving superior temporal consistency alongside enhanced visual quality.

\begin{figure*}[!ht]
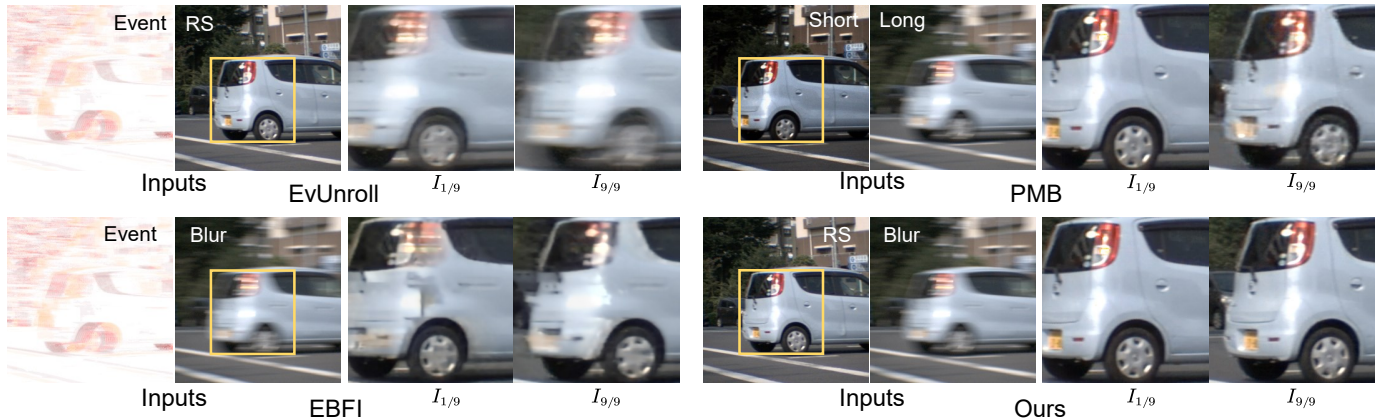

	\mfigure{1}{exp/compare_settings_realbr.pdf}
	\vspace{-5mm}
	\caption{
		\textbf{Qualitative comparison with competitive settings on realBR.} $I_t$ denotes the recovered latent frame at time $t$.
	}
	\label{fig:compare_with_competitive_setting_realbr}
	\vspace{-4mm}
\end{figure*}

\heading{Computational Complexity}
\tabref{compare_with_sota} also presents the computed complexity of all algorithms in terms of inference time, the amount of model parameters and computation FLOPs. Specifically, the FLOPs and running time were evaluated by recovering $9$ latent frames with a size of $256 \times 256$ on an NVIDIA Geforce RTX 3090. The results indicate that our method achieves a substantial lead in performance while maintaining moderate computational complexity. To enhance practical utility, we further introduce a lightweight variant, Ours-lite, which reduces channel widths and removes knowledge distillation and post-refinement, shrinking the model to 49\% of the original with RIFE-comparable parameters, while maintaining competitive performance on realBR and GOPRO-BR as shown in~\tabref{compare_with_sota} and ~\tabref{compare_with_sota_gopro}, demonstrating the scalability and practical value of our method.

\subsection{Comparison with Competitive Settings}
\label{comp_setting}
Recently, significant progress has been made in addressing image restoration tasks using multi-camera systems, demonstrating the effectiveness of multi-sensor and cross-modality strategies under complex visual conditions. To validate the advantages of our model and configuration, we compare it against SOTAs under competitive settings, including IFED~\cite{zhong2022bringing} with dual reversed RS views, EvUnroll~\cite{zhou2022evunroll} and EBFI~\cite{weng2023event} with aligned event cameras, and PMB~\cite{rengarajan2020photosequencing}, which fuses short- and long-exposure inputs.

Quantitative results are presented in~\tabref{compare_with_competitive_setting}, where our method consistently outperforms all competing setups on both synthetic and real datasets, corroborating the findings presented in \secref{comp_sota}. Additionally, we also adapt RIFE and IFED to the short-/long-exposure setting, denoted as RIFE$_{BS}$ and IFED$_{BS}$, respectively. Corresponding results further validate the superiority of our setting. Representative reconstruction results are visualized in \figref{compare_with_competitive_setting_gopro} and \figref{compare_with_competitive_setting_realbr}. Our approach not only delivers better visual quality compared to all cross-sensor configurations but also offers higher hardware efficiency than both EvUnroll and EBFI.

\begin{figure}[!tp]
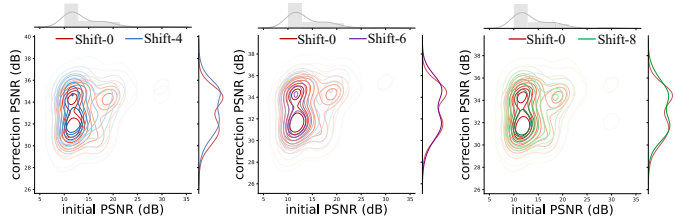

	\mfigure{1}{exp/misaligned_drawing.pdf}
	\vspace{-6mm}
	\caption{
		\textbf{PSNR distribution} of our method with one aligned (`Shift-$0$') and three spatially-misaligned views (`Shift-$4$', `Shift-$6$' and `Shift-$8$') under a selected sequence. The horizontal axis is initial PSNR computed by blur view and the first latent frame while the vertical axis denotes PSNR computed between corrected output and its ground truth.
	}
	\label{fig:misaligned_views}
	\vspace{-4mm}
\end{figure}

\begin{table}[!t]
	\centering
	\caption{\textbf{Quantitative comparisons under spatial misalignment and low-lit scenes}. `Shift-$n$' denotes misalignment with maximal offsets $n$ and `Noise-$m$' is an experiment conducted on synthesized low-light RS view under peak value $m$.}
	\label{tab:compare_with_shift}
	\resizebox{\linewidth}{!}
	{
		\begin{tabular}{rccccccccc}
			\toprule
			\multirow{2}{*}{Method} & \multicolumn{3}{c}{$\times$3} & \multicolumn{3}{c}{$\times$5} &  \multicolumn{3}{c}{$\times$9} \\ \cmidrule(lr){2-4}\cmidrule(lr){5-7}\cmidrule(lr){8-10}
			
			& PSNR     & SSIM     & LPIPS    & PSNR    & SSIM    & LPIPS & PSNR    & SSIM    & LPIPS   \\ \midrule
			
			Shift-$4$ & 31.02  & 0.9090 & 0.0806  &  31.21 &  0.9124 & 0.0792 & 31.32 & 0.9144 & 0.0789 \\
			
			Shift-$6$ &30.94  & 0.9077 & 0.0882 & 31.15 & 0.9120 & 0.0876 & 31.28 & 0.9144 & 0.0866\\		 	
			
			Shift-$8$ & 31.06 & 0.9090 & 0.0853 & 31.25 & 0.9126 & 0.0852 & 31.36 & 0.9149 & 0.0854 \\
			
			Noise-$300$ &30.53 &0.9030 &0.0901 &30.75 & 0.9067 & 0.0893 & 30.86& 0.9086&0.0894 \\
			
			Noise-$500$ &30.72 &0.9032 &0.0917 &30.92 &0.9067&0.0902 & 31.03& 0.9087&0.0898 \\
			
			Noise-$800$ &30.72 &0.9036 &0.0912 &30.93 &0.9074&0.0901 & 31.04& 0.9095&0.0898\\
			
			Ours &  30.97 & 0.9073 & 0.0695  & 31.19  &0.9105  &  0.0675 & 31.31 & 0.9122 & 0.0671  \\ \bottomrule
		\end{tabular}
	}
	\vspace{-2mm}
\end{table}

\begin{table}[!tp]
	\centering
	\caption{\textbf{Quantitative comparisons under temporal misalignment}. `T-Shift' denotes experimental results on temporally-misplaced RS-Blur views. 
	}
	\label{tab:temporal_shift}
	\resizebox{\linewidth}{!}
	{
		\begin{tabular}{rccccccccc}
			\toprule
			\multirow{2}{*}{Method} & \multicolumn{3}{c}{$\times$3} & \multicolumn{3}{c}{$\times$5} &  \multicolumn{3}{c}{$\times$7} \\ \cmidrule(lr){2-4}\cmidrule(lr){5-7}\cmidrule(lr){8-10}
			
			& PSNR     & SSIM     & LPIPS    & PSNR    & SSIM    & LPIPS & PSNR    & SSIM    & LPIPS   \\ \midrule
			
			T-Shift & 35.01  & 0.9738 & 0.0351  &  35.47 &  0.9759 & 0.0328 & 35.57 & 0.9768 & 0.0320 \\
			Ours &  35.10 & 0.9743 & 0.0343  & 35.50  &0.9763  &  0.0324 & 35.63 & 0.9770 & 0.0317  \\ \bottomrule
		\end{tabular}
	}
	\vspace{-3mm}
\end{table}

\subsection{Performance on Challenging Scenarios}
\label{sec:challenging_scenes}

\heading{Misaligned Views}
In~\secref{imaging_system}, we adopt several strategies to ensure accurate alignment between the blurred and RS views. Nevertheless, since the RS view mainly provides motion guidance and complementary local details, such guidance is known to tolerate a certain degree of misalignment~\cite{zhong2022bringing}. We first evaluate robustness to spatial misalignment by introducing random horizontal and vertical shifts with maximum offsets of 4, 6, and 8 pixels. As shown in~\tabref{compare_with_shift} and~\figref{misaligned_views}, the resulting performance drop is negligible. We then further examine temporal misalignment by synthesizing exposure-delayed RS data on GOPRO-BR, where the RS exposure start is delayed by $\delta$ frames, with $\delta \sim \mathcal{U}([5,15])$. ~\tabref{temporal_shift} shows that performance degrades gracefully under temporal offsets, indicating that temporally misaligned shutters can still provide effective motion cues and complementary global-to-local information to each other. Notably, we retrained our model under misaligned views following original training strategy.

\begin{figure}[!tp]
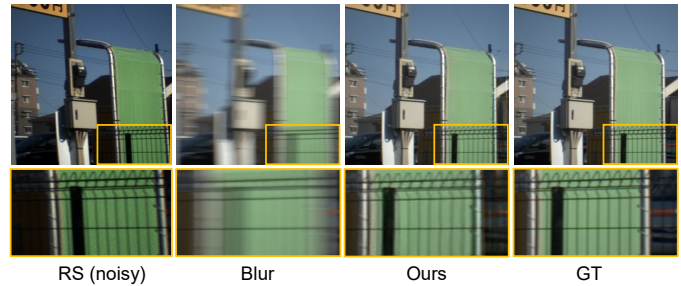

	\mfigure{1}{exp/low_lit_comparison_v2.pdf}
	\vspace{-6mm}
	\caption{
		\textbf{Visual results} of our proposed method under low-lit scenes with peak value $500$. Due to short exposure time of each rows in RS view, it suffers from obvious noise. But our setting is still capable of dealing with this challenge.
	}
	\label{fig:low_lit_scenes}
	\vspace{-3mm}
\end{figure}

\begin{table}[!tp]
	\centering
	\caption{\textbf{Quantitative results of our model on the Blurred-RS and Occluded-RS testsets.} The results indicate the effectiveness of our method under these challenging conditions.}
	\label{tab:occlusion_blur_in_rs}
	\resizebox{\linewidth}{!}
	{
		\begin{tabular}{lccccccccc}
			\toprule
			\multirow{2}{*}{Data} & \multicolumn{3}{c}{$\times$3} & \multicolumn{3}{c}{$\times$5} &  \multicolumn{3}{c}{$\times$9} \\ \cmidrule(lr){2-4}\cmidrule(lr){5-7}\cmidrule(lr){8-10}
			
			& PSNR     & SSIM     & LPIPS    & PSNR    & SSIM    & LPIPS & PSNR    & SSIM    & LPIPS   \\ \midrule
			Blurred-RS Testset &  30.46 & 0.9034 & 0.0633  & 30.64  &0.9062  &  0.0621 & 30.74 & 0.9077 & 0.0617  \\
			Occluded-RS Testset &  30.96 & 0.9031 & 0.0678  & 31.18  &0.9068  &  0.0665 & 31.30 & 0.9088 & 0.0664  \\ \bottomrule
		\end{tabular}
	}
	\vspace{-4mm}
\end{table}

\heading{Low-lit Scenes}
Following conventional practice, we select appropriate row exposure times to avoid saturation in the GS view and reduce motion blur in the RS view. However, under low-light conditions, the RS observation is prone to noise. To investigate the impact of such noise on our method, we simulate low-light RS captures by applying random gamma correction and Poisson noise to clean RS images, following the approach of~\cite{lv2018mbllen, lore2017llnet}. We vary the peak values to simulate different noise intensities. The quantitative results in~\tabref{compare_with_shift} show that low-light scenes lead to a PSNR drop of approximately 0.45 dB, yet our method still outperforms the second-best approach reported in~\tabref{compare_with_sota}. Visual results are provided in~\figref{low_lit_scenes}.

\begin{figure}[!tp]
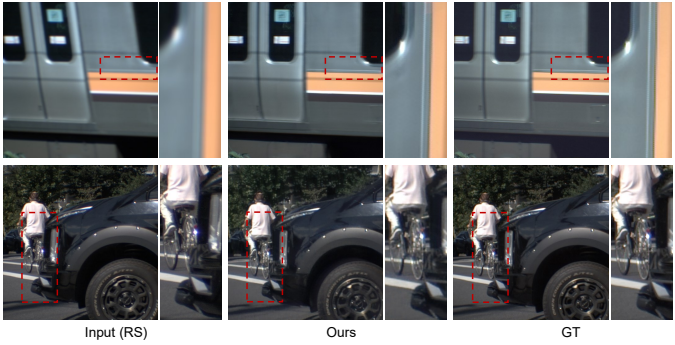

	\mfigure{1}{exp/occlusion_blur_in_rs.pdf}
	\vspace{-6mm}
	\caption{
		\textbf{Qualitative results on the Blurred-RS and Occluded-RS testsets.} The first row shows RS inputs with motion blur, and the second row shows occlusion cases.
	}
	\label{fig:occlusion_blur_in_rs}
	\vspace{-2mm}
\end{figure}

\heading{Occluded and Blurred RS View}
In real-world scenarios, RS views may contain motion blur and complex occlusions caused by dynamic objects and camera motion. To evaluate these challenging conditions, we evaluate our model on targeted Blurred-RS Testset and an Occluded-RS Testset. Quantitative and qualitative results (\tabref{occlusion_blur_in_rs} and \figref{occlusion_blur_in_rs}) show that our method maintains reliable performance under these situations, benefiting from the complementary information between RS and blur observations.

\begin{table}[!tp]
	\centering
	\setlength\tabcolsep{4pt}
	\caption{\textbf{Architecture ablation on our realBR dataset.} TPE denotes temporal positional encoding for RS review and SAA is the shutter alignment and aggregation module. Performance is evaluated using mean PSNR/SSIM/LPIPS/tOF.}
	\resizebox{\columnwidth}{!}{
		\begin{tabular}{ccccccc} 
			\toprule
			Variants & TPE & SAA & Dual-stream & Distillation  & Prompt & Metrics \\
			\midrule
			{V1} &  &\checkmark  & \checkmark &   &  & 31.06/0.9104/0.0690/0.70 \\
			{V2} & \checkmark &  & \checkmark &     &  & 27.96/0.8645/0.1442/0.66 \\
			{V3} &\checkmark  & \checkmark &  &    &  & 30.33/0.9013/0.0929/0.57 \\
			{V4} & \checkmark  & \checkmark  & \checkmark &   &  & 31.15/0.9120/0.0678/0.45\\
			{V5} & \checkmark  & \checkmark  & \checkmark &   & \checkmark & 31.26/0.9097/0.0828/0.37  \\
			{V6} & \checkmark  & \checkmark  & \checkmark &  \checkmark &  & 31.25/0.9096/0.0621/0.31  \\
			\midrule
			Ours & \checkmark  & \checkmark   & \checkmark  &\checkmark  & \checkmark & 31.31/0.9122/0.0671/0.23\\
			\bottomrule
		\end{tabular}
	}
	\label{tab:structure_ablation}
	\vspace{-2mm}
\end{table}

\begin{figure}[!tp]
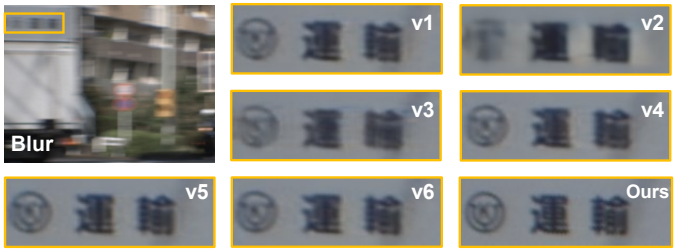

	\mfigure{1}{exp/structure_ablation_v2.pdf}
	\vspace{-6mm}
	\caption{
		\textbf{Visual results} of ablation study on model structure design. Best viewed in zoom.
	}
	\label{fig:structure_ablation}
	\vspace{-4mm}
\end{figure}

\subsection{Ablation Study}

\heading{Effects of Components for Moment Reenacting}
To investigate the contribution of each component to imaging moment reaction, we perform an ablation study by incrementally introducing temporal positional encoding (TPE), shutter alignment and aggregation (SAA), dual-stream architecture, self-prompt learning, and an adaptive distillation strategy. The corresponding results are presented in~\tabref{structure_ablation} and~\figref{structure_ablation}. Notably, a single-stream approach that simply concatenates the blur and RS views performs suboptimally, as it fails to harness the complementary information embedded in the cross-shutter configuration. In contrast, our dual-branch design, connected via the SAA module, effectively captures the distinct yet synergistic roles of the RS and blur views. This mutual incentive mechanism enhances the model’s ability to resolve the inherent ambiguities of the task. The integration of TPE alone brings an additional improvement of approximately 0.1 dB. Building on this (v4), incorporating knowledge distillation and self-prompt learning further elevates reconstruction performance. The distillation mechanism helps the student model better capture subtle motion cues, while the motion-residue prompt dynamically refines the warped frames by accounting for spatially varying confidence levels. To further analyze the contribution of SAA, we visualize the estimated $F_{B \rightarrow R}$ and $F_{R \rightarrow B}$ (\figref{biflow_vis}). These flows establish cross-view correspondences for feature warping, enabling aligned complementary cues and improving the robustness of cross-shutter feature interaction.

\begin{figure}[!tp]
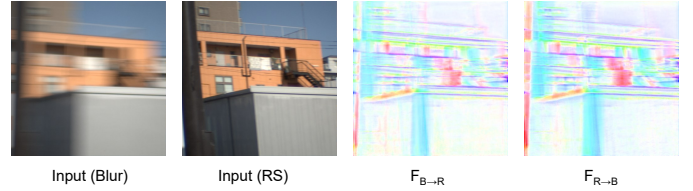

	\mfigure{1}{exp/biflow_vis.pdf}
	\vspace{-6mm}
	\caption{
		\textbf{Visualization} of bidirectional flow fileds in SAA.
	}
	\label{fig:biflow_vis}
	\vspace{-4mm}
\end{figure}

\begin{table}[!th]
	\centering
	\setlength\tabcolsep{4pt}
	\caption{\textbf{Effects of latent motion distillation.} We compare our region-adaptive distillation with other strategies.}
	\resizebox{\columnwidth}{!}{
		\begin{tabular}{lccc} 
			\toprule
			Experiments & PSNR  & SSIM  & LPIPS \\
			\midrule
			(a) Without distillation & 31.26 & 0.9097  & 0.0828       \\
			(b) Without distillation mask & 31.00 & 0.9064 & 0.0668   \\
			(c) Without the mask for dynamic areas ($\textbf{M}_d$) &31.16  & 0.9116 & 0.0771  \\
			(d) Without the mask for boundaries ($\textbf{M}_b$) & 31.13  &  0.9113 &0.0686 \\
			(e) Without the mask for low confidence($\textbf{M}_e$) & 31.22  & 0.9117  &0.0715     \\
			\midrule
			(f) $\textbf{M}_d + \textbf{M}_b + \textbf{M}_e$ (Ours) & 31.31  &  0.9122  & 0.0671  \\
			\bottomrule
		\end{tabular}
	}
	\label{tab:distillation_ablation}
	\vspace{-3mm}
\end{table}

\heading{Effects of Adaptive Motion Distillation}
To further verify the effectiveness of our adaptive motion distillation, we conduct an ablation study on various distillation operations. Specifically, we examine the impact of removing: the entire distillation process, all masking mechanisms, and individual masks targeting dynamic regions, object boundaries, and low-confidence estimates. The experimental results are summarized in ~\tabref{distillation_ablation}. As shown in (a) and (b), direct incorporation of privileged distillation incurs a 0.26 dB performance drop. This supports the finding in~\cite{li2023multi} that, without sufficient regularization, the teacher model tends to over-utilize privileged information, which impairs its generalization ability on the test set. Thus, we introduce three types of mask, $\textbf{M}_d$, $\textbf{M}_b$ and $\textbf{M}_e$, actively addressing the distillation on highly dynamic areas, object boundaries and low-confidence estimates. The results confirm that combining all three masks leads to the best scores.

\begin{figure}[!tp]
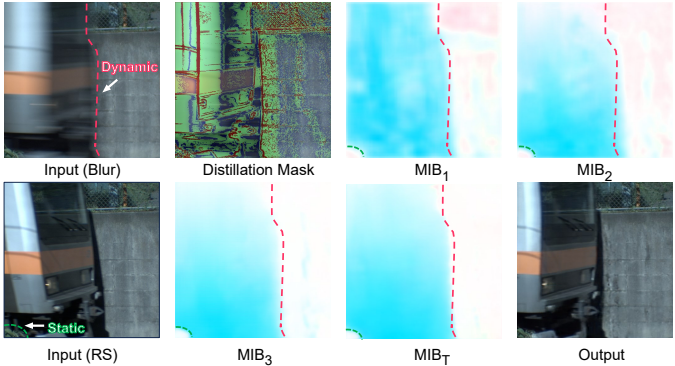

	\mfigure{1}{exp/distill_visualization.pdf}
	\vspace{-6mm}
	\caption{
		\textbf{Visualization} of intermediate flows from student module: MIB$_1$, MIB$_2$, MIB$_3$, teacher module: MIB$_T$ and our distillation mask.
	}
	\label{fig:distill_visualization}
	\vspace{-3mm}
\end{figure}

Besides, we visualize the predicted intermediate flows from both the student and teacher modules, as well as the proposed distillation mask, in~\figref{distill_visualization}. Our MIBs are capable of reconstructing latent motions by accurately distinguishing between static and dynamic objects, and the designed distillation mask effectively mitigates the overuse of privileged knowledge.

\begin{table}[!tp]
	\centering
	\setlength\tabcolsep{4pt}
	\caption{\textbf{Effects of self-prompted reconstruction.} We compare our motion-residue prompt with other strategies.}
	\resizebox{\columnwidth}{!}{
		\begin{tabular}{lccc} 
			\toprule
			Experiments & PSNR  & SSIM & LPIPS  \\
			\midrule
			(a) Without  prompt &31.25  &  0.9096 & 0.0621       \\
			(b) Learnable prompt & 31.16 & 0.9085 &  0.0696  \\
			(c) Feature-level flow difference       &31.06  & 0.9100 &  0.0671   \\
			(d) GT-informed flow difference &  22.66 &  0.8010 & 0.2891    \\
			(e) Concatenation of context and flow feature & 30.79  & 0.9051 & 0.0873 \\
			\midrule
			(f) Image-level flow difference (Ours) &  31.31  &  0.9122  & 0.0671  \\
			\bottomrule
		\end{tabular}
	}
	\label{tab:prompt_ablation}
	\vspace{-1mm}
\end{table}

\begin{figure}[!tp]
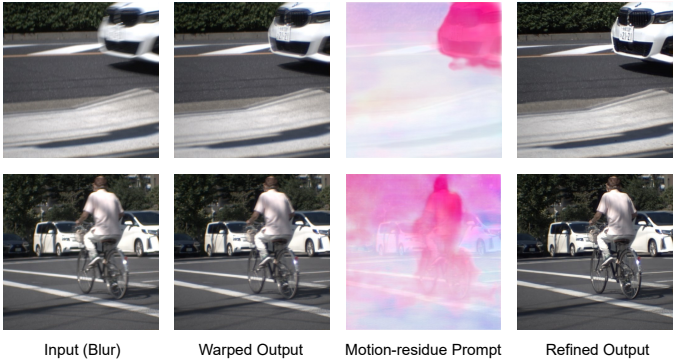

	\mfigure{1}{exp/prompt_visualization.pdf}
	\vspace{-2mm}
	\caption{
		\textbf{Visualization} of warped intermediate frames and their motion-residue prompts. Best viewed in zoom.
	}
	\label{fig:prompt_visualization}
	\vspace{-2mm}
\end{figure}

\heading{Effects of Motion-residue Prompt}
In this section, several prompt strategies have been explored, including GT-informed difference computation~\cite{wang2024selfpromer} and learnable prompts~\cite{liang2023iterative}. We also evaluate prompts derived from feature-level difference maps and concatenation of context and flow features, as proposed in~\cite{wang2024selfpromer}. The results, presented in ~\tabref{prompt_ablation}, reveal that the learnable prompt with random initialization performs reasonably well but falls short in terms of perceptual quality. Prompts based on feature-level flow differences or context-flow concatenation yield slightly inferior results. However, all these variants underperform compared to the baseline without any prompt, indicating that not all prompts lead to performance improvements. In particular, the GT-informed flow difference performs significantly worse across all metrics due to the continuously iterative inference from zero difference map causes convergence issues. In contrast, our proposed prompt achieves the best performance, demonstrating its effectiveness in guiding reconstruction with reliable and generalizable motion cues.

The warped intermediate frames and their corresponding motion-residue prompts are visualized in~\figref{prompt_visualization}. Our self-prompter effectively identifies regions with complex motion, exactly where geometric structures and local textures require further enhancement. This enables the frame reconstruction module to perform targeted and precise refinement on the warped outputs.

\subsection{Third-party Evaluation}
To comprehensively evaluate the generalization and adaptability of our method to various capture settings and real-world scenes, we additionally collected a third-party testset that includes a wide range of motions and objects. Specifically, we made the following changes:(1) We use another types of sensors to capture RS frames (EO-1312LE) and blurred frames (BFLY-U3-23S6C-C), both at a resolution of $512 \times 512$. (2) All input videos are captured at 50 fps with no deadtime. (3) The exposure time is adjusted to 20 ms to generate different intensities of degradation.

\begin{figure}[!tp]
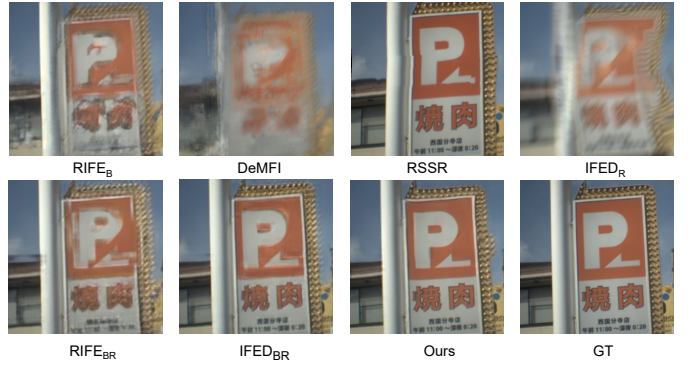

	\mfigure{1}{exp/thirdparty_test.pdf}
	\vspace{-4mm}
	\caption{
		\textbf{Qualitative comparison} with representative SOTAs on third-party testset.
	}
	\label{fig:thirdparty_test}
	\vspace{-2mm}
\end{figure}

\begin{table}[!tp]
	\centering
	\caption{\textbf{Quantitative comparison} with representative SOTAs on third-party testset.}
	\label{tab:thirdparty_test}
	\resizebox{\linewidth}{!}
	{
		\begin{tabular}{rccccccc}
			\toprule
			\multirow{2}{*}{Method}& \multirow{2}{*}{Input} & \multicolumn{3}{c}{$\times$3} & \multicolumn{3}{c}{$\times$9} \\ \cmidrule(lr){3-5}\cmidrule(lr){6-8}
			
			& & PSNR     & SSIM     & LPIPS    & PSNR    & SSIM    & LPIPS    \\ \midrule

			RIFE$_{B}$~\cite{huang2022real} &  \emph{2}$\cdot$\emph{B}  & 26.31 & 0.8679 & 0.1630  & 26.52 & 0.8710 & 0.1633    \\  \midrule

			DeMFI~\cite{oh2022demfi} & \emph{4}$\cdot$\emph{B} & 25.31  & 0.8471 & 0.2180 & 24.94  & 0.8428 & 0.2249   \\  \midrule
			
			RSSR ~\cite{fan2021inverting}& \multirow{2}{*}{\emph{2}$\cdot$\emph{R}}  & 25.26 &  0.8292 & 0.0833 & 24.78 & 0.8170  & 0.0933   \\ 			
			IFED$_{R}$~\cite{zhong2022bringing}& & 22.70 & 0.7979  & 0.2452 & 22.98 & 0.8057  &0.2407   \\      
			\midrule
			
			RIFE$_{BR}$~\cite{huang2022real} &  \multirow{3}{*}{\emph{B}$\cdot$\emph{R}}  & 29.26 & 0.9054 & 0.1139   & 29.58 & 0.9112 & 0.1022    \\
			
			IFED$_{BR}$~\cite{zhong2022bringing} &   & 29.06  & 0.8999 & 0.0705 & 28.98 & 0.8997 &  0.0675    \\
			
			Ours\quad\,\,\,\,\, &  & 29.56 & 0.9107  & 0.0882  & 29.88 & 0.9159 & 0.0844    \\ \bottomrule
		\end{tabular}
	}
	\vspace{-2mm}
\end{table}

The experimental results in~\tabref{thirdparty_test} and ~\figref{thirdparty_test} are obtained by directly inferring on the third-party testset using models pre-trained on realBR. It is clear that our model obtain the best performance perceptually and numerically.

\begin{figure*}[!tp]
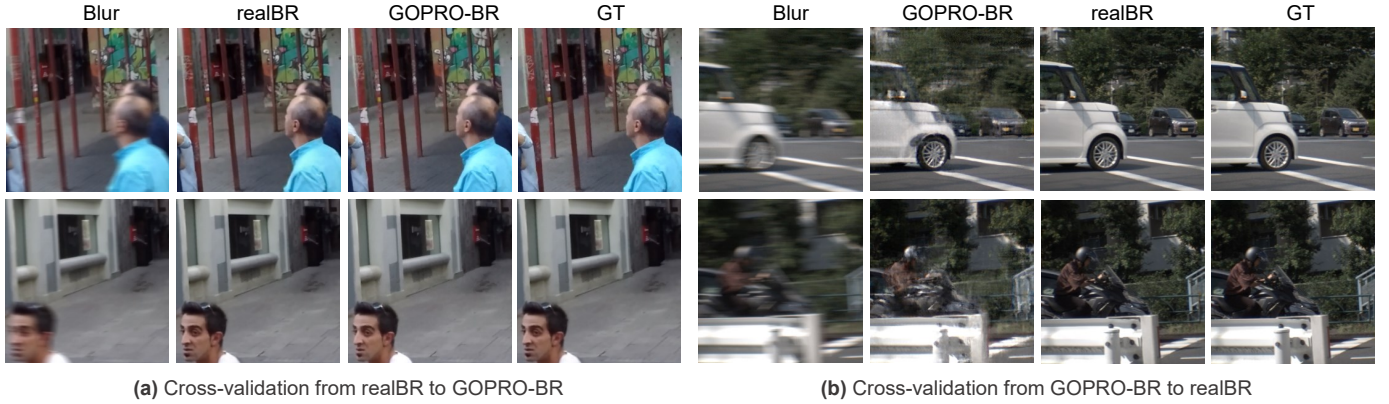

	\mfigure{1}{exp/cross_validation.pdf}
	\vspace{-4mm}
	\caption{
		\textbf{Cross-validation between realBR and GOPRO-BR.} (a) Visualized samples from test set of GOPRO-BR by using our model trained on realBR and GOPRO-BR. (b) Visualized samples from test set of realBR by using our model trained on GOPRO-BR and realBR. Best viewed in zoom.
	}
	\label{fig:cross_validation}
	\vspace{-4mm}
\end{figure*}

\begin{table}[!hp]
	\centering
	\caption{\textbf{Dataseet cross-validation} in PSNR/SSIM/LPIPS when recovering 7 latent frames. The first column denotes bottom scores computed with RS-blur input and training dataset for our model. The first row is data used for testing.}
	\label{tab:cross_validation} 
	\resizebox{\linewidth}{!}
	{
		\begin{tabular}{rccc}
			\toprule
			
			& GOPRO-BR     & realBR     & Third-party      \\   \cmidrule(lr){2-4}
			
			RS  & 21.04/0.7706/0.0914  & 19.05/0.6692/0.2145  & 25.43/0.8297/0.0827    \\  \cmidrule(lr){2-4}
			
			Blur & 23.31/0.8293/0.2475 & 21.02/0.7668/0.2961  & 24.41/0.8542/0.1876  \\ \toprule
			
			GORPO-BR&  35.63/0.9770/0.0317 & 21.51/0.7686/0.1730   & 24.06/0.8473/0.1123  \\ \cmidrule(lr){2-4}
			
			realBR& 27.35/0.9132/0.1384 &  31.34/0.9123/0.0655  & 29.93/0.9167/0.0843   \\ \bottomrule
		\end{tabular}
	}
	\vspace{-2mm}
\end{table}

\subsection{Dataset Cross-Validation}
To evaluate the benefits of using real-world data for training, we conducted cross-validation among GOPRO-BR, realBR, and a third-party testset. As shown in~\figref{cross_validation}a, the model trained on realBR performs well on the GOPRO-BR testset. In contrast,~\figref{cross_validation}b shows that the model trained on GOPRO-BR struggles on the realBR testset, producing noticeable artifacts. This comparison highlights that training on real-world data significantly improves generalization, whereas models trained on synthetic data fail to handle real-world scenarios effectively. 

We also present a quantitative comparison for cross-dataset validation, as shown in~\tabref{cross_validation}. The bottom scores for each dataset are computed using the original inputs, RS or blurry frames and their corresponding sharp ground truth (e.g., the first latent frame). It can be observed that the model trained on realBR consistently achieves performance gains on both synthetic and third-party datasets. However, the model trained on GOPRO-BR often yields results that are comparable to or worse than the bottom scores, indicating its inability to reconstruct latent frames effectively and, in some cases, even degrading original quality of testsets.

\begin{figure}[!tp]
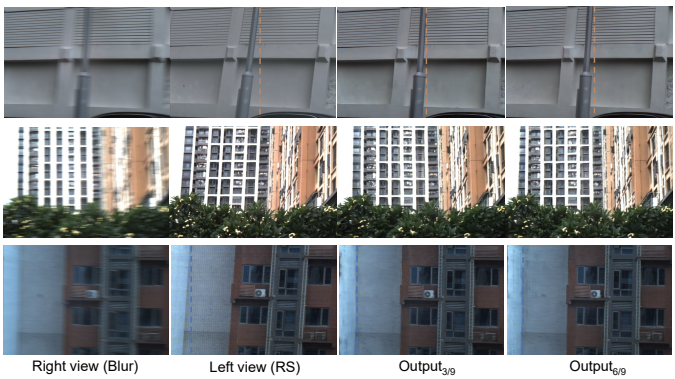

	\mfigure{1}{exp/stereo_setup.pdf}
	\vspace{-2mm}
	\caption{
		\textbf{Visual results} of our model ($d_{\text{up}}=20$) on real stereo blur-RS testset. Output$_t$ denotes the reconstructed latent frame at time instant $t$.
	}
	\label{fig:stereo_setup}
	\vspace{-3mm}
\end{figure}

\begin{table}[!tp]
	\centering
	\caption{\textbf{Quantitative results of our model under stereo configuration.} We evaluate the performance of reconstructing latent frame sequence with lengths of $3$ and $9$.}
	\label{tab:stereo_setup}
	\resizebox{\linewidth}{!}
	{
		\begin{tabular}{lccccccc}
			\toprule
			\multirow{2}{*}{Model}& \multirow{2}{*}{Setup} & \multicolumn{3}{c}{$\times$3} & \multicolumn{3}{c}{$\times$9} \\ \cmidrule(lr){3-5}\cmidrule(lr){6-8}
			
			& & PSNR     & SSIM     & LPIPS    & PSNR    & SSIM    & LPIPS    \\ \midrule
			
			Ours-DualBR & \emph{B}$\cdot$\emph{R}  &  30.97 & 0.9073 & 0.0695  & 31.31 & 0.9122 & 0.0671 \\ \midrule
			
			Ours-StereoBR & $d_{\text{up}}=20$  & 30.26 & 0.9003 & 0.0961   & 30.58 & 0.9062 & 0.0950    \\
			
			Ours-StereoBR &  $d_{\text{up}}=25$ & 29.98  & 0.8965 & 0.0992 &30.38 & 0.9028 &  0.0981    \\
			Ours-StereoBR  & $d_{\text{up}}=30$& 28.85 & 0.8722  & 0.1084  & 29.15 & 0.8805 & 0.1055    \\
			Ours-StereoBR  & $d_{\text{up}}=35$ & 28.22 & 0.8675  & 0.1153  & 28.43 & 0.8739 & 0.1131    \\ \midrule
			RIFE-StereoBR & $d_{\text{up}}=25$  & 26.58 & 0.8503 &  0.1755  & 26.74 & 0.8544 & 0.1751   \\		
			IFED-StereoBR &  $d_{\text{up}}=25$ & 26.33 & 0.8254 & 0.1383 & 26.57 & 0.8326 &  0.1340    \\ \midrule	
			RIFE-StereoB & $d_{\text{up}}=25$  & 21.98 & 0.7757 &  0.3178  & 22.43 & 0.7880 & 0.3134   \\		
			IFED-StereoB &  $d_{\text{up}}=25$ &  21.63 & 0.7643 & 0.2705 & 22.06 & 0.7776 &  0.2647    \\
			Ours-StereoB &  $d_{\text{up}}=25$ & 21.72 & 0.7642 & 0.2436 & 22.14 & 0.7769 &  0.2407    \\ \midrule
			RIFE-StereoR & $d_{\text{up}}=25$  & 21.75 & 0.7593 & 0.2595   & 22.35 & 0.7780 & 0.2343   \\		
			IFED-StereoR &  $d_{\text{up}}=25$ & 21.26 & 0.7444 & 0.2271 & 21.76 & 0.7620 &  0.2055    \\
			Ours-StereoR &  $d_{\text{up}}=25$ & 21.98 & 0.7516 & 0.1697 & 22.65 & 0.7719 &  0.1492    \\
			\bottomrule
		\end{tabular}
	}
	\vspace{-2mm}
\end{table}

\begin{figure*}[!tp]
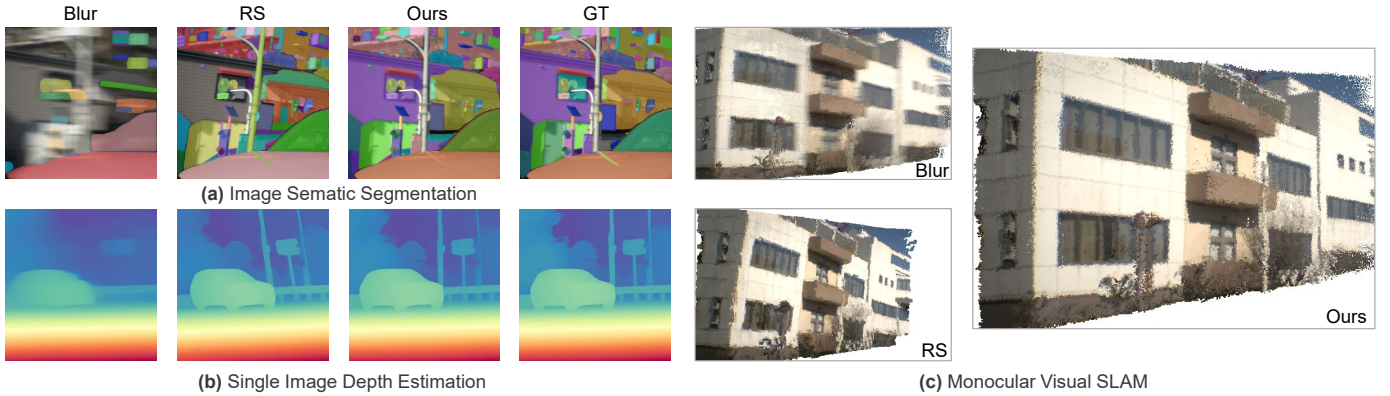

	\mfigure{1}{exp/downstream_application_v2.pdf}
	\vspace{-3mm}
	\caption{
		\textbf{Downstream applications.} We conduct segmentation~\cite{kirillov2023segment}, depth estimation~\cite{yang2024depth} and SLAM~\cite{liu2024slam3r} on raw RS, blur frames along with processed latent videos using our model.
	}
	\label{fig:downstream_application}
	\vspace{-2mm}
\end{figure*}

\subsection{Dual-to-Stereo Extension}

The current implementation relies on a beamsplitter to precisely align the two different shutters. Although we demonstrated in~\secref{challenging_scenes} that the dual-shutter setup exhibits a degree of tolerance to misalignment, perfect coaxial imaging mechanism may not always be satisfied in real-world scenarios, particularly on compact mobile devices. Therefore, we make a step forward to extend the proposed dual-shutter framework to a more general configuration: stereo GS-RS cameras with a narrow baseline, a setup commonly found in smartphones and tablets.

Using the realBR dataset, we first follow the method in~\cite{watson2020learning,guo2024stereo,cun2018depth} to synthesize stereo counterparts of RS videos and combine them with the original blurred videos to form stereo blur-RS pairs. After investigating the imaging parameters of typical mobile devices, we determine the upper bound of maximum disparity $d_{\text{up}}$, in accordance with the scope of baselines ($5\text{-}20\mathrm{mm}$) and focal lengths ($2\text{-}5\mathrm{mm}$). Considering the minimum depth range in real urban scenes ($2\text{-}4\mathrm{m}$), we set the $d_{\text{up}}$ to 20, 25, 30 and 35 to synthesize the corresponding stereo datasets. We keep the model architecture unchanged and retrain it on these stereo datasets following the same training and testing protocols. ~\tabref{stereo_setup} summarizes the experimental results under this setting. Although the stereo solutions incur performance drops compared with Ours-DualBR, we can observe that they outperform the existing methods in~\tabref{compare_with_sota} and ~\tabref{compare_with_competitive_setting}. We additionally introduce two sets of experiments: (1) adapting RIFE and IFED to the stereo Blur–RS configuration (stereoBR); and (2) introducing new baselines that use pure stereo Blur inputs (stereoB) and pure stereo RS inputs (stereoR), respectively, and retraining our model, RIFE, and IFED under these settings. Overall, the results suggest the advantage of stereoBR over single-modality stereo inputs and the robustness of our model to stereo misalignment.

To comprehensively demonstrate the effectiveness and generalizability of our stereo extension, we collect real-world Blur–RS pairs using a tailored stereo camera and construct a stereoBR-testset consisting of 13 diverse urban scenes. The stereo imaging system is directly derived from the setup described in~\secref{imaging_system}, with several modifications. Specifically, the beam-splitter group and the high-speed camera are removed, and the RS camera and GS camera are horizontally aligned. All other hardware configurations remain the same as those described in~\secref{imaging_system}. As illustrated in~\figref{stereo_setup}, our model successfully recovers accurate geometry and fine details from stereo Blur-RS degradation. Notably, we also provide low-lit scenes to explore the robustness of our solution under noises. More visual results obtained by pre-trained model with different $d_{\text{up}}$ are available on \href{https://jixiang2016.github.io/dualBR_site/}{\textit{Project Page}}. The stereo extension ensures relatively high performance while reducing hardware requirements, significantly broadening the application scenarios and allowing us to choose the most suitable strategy based on specific needs.

\begin{table}[!tp]
	\centering
	\setlength{\tabcolsep}{4pt}
	\caption{\textbf{Quantitative comparison on single-image deblurring task} in terms of PSNR/SSIM/LPIPS}
	\vspace{-1mm}
	\resizebox{1\columnwidth}{!}{
		\begin{tabular}{ccccc} 
			\toprule
			 \makecell[c]{EVSSM {\cite{kong2025efficient}}} & \makecell[c]{AdaRevD  {\cite{mao2024adarevd}}} & \makecell[c]{FFTformer  {\cite{kong2023efficient}}} & \makecell[c]{DeblurDiff  {\cite{kong2025deblurdiff}}} & Ours \\
			\midrule
			21.14/0.7589/0.3524 & 27.01/0.8619/0.1937 & 26.36/0.8119/0.2775 & 21.08/0.7460/0.2127 & 28.42/0.8768/0.1106 \\
			\bottomrule
		\end{tabular}
	}
	\label{tab:single_image_deblurring}
	\vspace{-3mm}
\end{table}

\subsection{Comparison on Single-image Deblurring}

We compare our method with SOTA single-image deblurring approaches in~\tabref{single_image_deblurring}. As shown, our method consistently outperforms these methods. This improvement mainly comes from the complementary nature of the dual-view inputs: the RS image provides local details, while the blur image preserves globally integrated scene information. Their combination effectively reduces the inherent ambiguity of single-image deblurring. For fairness, we adopt a unified training setting by training all methods for 500 epochs.

\subsection{Downstream Application}
Our solution significantly enhances the visual quality of input videos by reconstructing latent frames affected by motion degradation. This improvement benefits a wide range of downstream tasks in real-world scenarios, such as depth estimation~\cite{yang2024depth, ji2018dense}, semantic segmentation~\cite{kirillov2023segment}, and 3D reconstruction~\cite{ye2020drm, liu2024slam3r}. To demonstrate these advantages, we conduct experiments on single-image depth estimation, semantic segmentation, and monocular visual SLAM using raw RS, motion-blurred frames, and our recovered latent videos. The results in~\figref{downstream_application} show that motion degradation significantly impairs the accurate perception of scene semantics and structure, leading to erroneous depth predictions and object recognition. In contrast, our model effectively corrects motion-induced distortions and accurately reconstructs latent motion, making downstream visual algorithms more robust in complex and dynamic environments.

\section{Conclusion}
\label{sec:conclusion}

In this paper, we present a unified framework for motion degradation inversion by jointly addressing blur decomposition and RS temporal super-resolution within a novel dual-shutter imaging setting. Leveraging complementary characteristics of global shutter (GS) and rolling shutter (RS) views, our approach reenacts the imaging moment under motion-induced degradations. To support this framework, we design a triaxial imaging system capable of capturing synchronized Blur-RS pairs along with high-speed ground-truth sequences, and construct a high-quality real-world dataset to enable comprehensive training and evaluation beyond synthetic data. Furthermore, we extend the dual-shutter setup to a narrow-baseline stereo Blur-RS configuration, offering flexible performance-cost trade-offs for practical deployment. On the algorithmic side, we propose a dual-stream motion interpretation network that jointly captures contextual semantics and temporal order, coupled with a self-prompted reconstruction module that refines sharp frames based on motion residuals. Extensive experiments on both real and synthetic datasets validate not only the effectiveness, generalizability, and superiority of our method, but also its practical applicability in real-world scenarios.

	\clearpage
	\small
	\bibliographystyle{IEEEtran}
	\bibliography{reference}

	\appendices
	\section{Implementation Details}

\subsection{Extended Discussion on Motion-residue Prompt}

In this section, we provide an extended discussion of our prompt-based frame reconstruction module, which is omitted from the main paper for conciseness.

The warping–refinement paradigm has achieved notable success in video reconstruction under motion degradation~\cite{oh2022demfi,liu2020deep,zhong2022animation,fan2021inverting,fan2022context}. Existing studies primarily focus on improving motion estimation, for example by designing more efficient building blocks~\cite{fan2023joint,purohit2019bringing}, directly estimating intermediate flow~\cite{zhong2022bringing,shang2023self}, or introducing attention mechanisms~\cite{zhong2021towards,ji2023rethinking}, while the refinement stage remains relatively underexplored. Most methods adopt a U-shaped architecture for refinement to capture multi-scale context, but typically treat all spatial regions uniformly, without accounting for varying reliability in warped frames. This lack of region-aware guidance can propagate warping-induced distortions, particularly in areas with complex motion or occlusions, ultimately degrading restoration quality.

On the other hand, recent work has explored prompt-based guidance to address spatially varying degradations. Liu \etal~\cite{liu2023pre} introduce NLP-based prompt learning inspired by human visual correction mechanisms, but such approaches are less applicable to vision-only settings. Subsequent studies investigate text-free prompting for visual tasks~\cite{herzig2024promptonomyvit,gan2023decorate}. For example, PromptonomyViT~\cite{herzig2024promptonomyvit} leverages diverse prompts such as depth and segmentation maps, while Wang \etal~\cite{wang2024selfpromer} propose a self-prompt dehazing Transformer based on depth difference maps between hazy images and corresponding GTs. Nonetheless, this method is accompanied by notable drawbacks: (1) the reliance on external depth estimation models can significantly increase computational complexity and potentially degrade performance, and (2) the absence of ground-truth during inference make the computation of difference maps non-trivial. Furthermore, the proposed continuously iterative approximation from zero initialization will be easily caught into local minima. These limitations further motivate the motion-residue prompted design introduced in the main paper.

\begin{table}[!t]
	\centering
	\caption{\textbf{Specifications of our triaxial imaging system}. The deadtime between two adjacent high speed frames is extremely short and thus can be ignored.
	}

	\setlength\tabcolsep{4pt}	
	\resizebox{0.9\linewidth}{!}
	{
		\begin{tabular}{lccc}
			\toprule
			\textbf{Device} & \textbf{RS camera} & \textbf{GS camera} & \textbf{HS camera}\\
			\midrule
			{Resolution} & $800\!\times\!800$  & $800\!\times\!800$ & $800\!\times\!800$\\
			{Frame rate} & 20 fps & 20 fps & 500 fps\\  
			{Exp. per Row} & 2 ms  & 18 ms & 2 ms\\ 
			{Delay. per Row} & 20 $\mu$s  & 0 $\mu$s & 0 $\mu$s\\ 
			{Exp. per Frame} & 18 ms  & 18 ms & 2 ms\\
			{Deadtime} & 32 ms  & 32 ms  & 0 ms\\ 
			\bottomrule
		\end{tabular}
	}
	\label{tab:imaging_system}
\end{table}

\begin{figure}[!t]
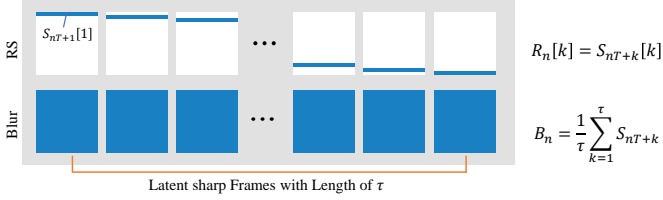

	\centering
	\mfigure{1}{supp/synthetic_process.pdf}
	\vspace{-2mm}
	\caption{
		\textbf{Synthetic method}. The notation $R[k]$ denotes extracting the k-th row from frame $R$. $n$ is the index of blur or RS view. $T$ is the number of latent frames that correspond to exposure and deadtime.
	}
	\label{fig:synthetic_process}
	\vspace{-2mm}
\end{figure}

\subsection{Construction of Imaging system}

We provide detailed specifications of the triaxial imaging system used in our data acquisition.

\tabref{imaging_system} summarizes the specifications of the RS, GS, and HS cameras in our triaxial imaging system, which share the same spatial resolution but differ in temporal sampling. Although the RS and GS cameras have the same frame exposure time (18 ms), the rolling shutter mechanism exposes rows sequentially with a 2 ms per row and a 20 $\mu$s inter-row delay, whereas the global shutter exposes all rows simultaneously. In contrast, the HS camera operates at 500 fps with a short 2 ms exposure and negligible deadtime, providing dense temporal sampling and ground truth.

\begin{figure*}[!h]
	\centering		
	\includegraphics[width=1\linewidth]{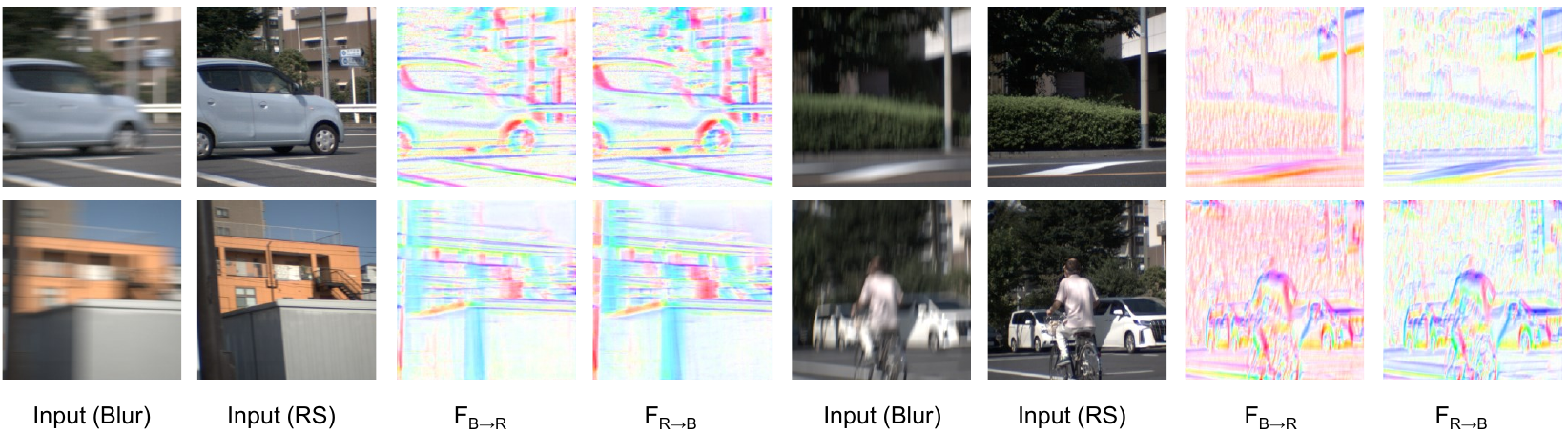}
	\caption{
		\textbf{Qualitative visualization of bidirectional displacement fields.} Given blur and RS inputs, we show the estimated forward and backward flows $F_{B \rightarrow R}$ and $F_{R \rightarrow B}$.These flows establish cross-view correspondences for feature warping in the SAA module, enabling effective alignment between the two views.
	}
	\label{fig:biflow_vis}
\end{figure*}

The HS camera (BITRAN CS-700C) is equipped with a SONY IMX426 sensor and air-forced cooling, resulting in exceptionally low noise. The IMX426 achieves an effective pixel pitch of $9\,\mu\mathrm{m} \times 9\,\mu\mathrm{m}$ by integrating four $4.5\,\mu\mathrm{m} \times 4.5\,\mu\mathrm{m}$ subpixels, while active cooling maintains the sensor at $0^\circ\mathrm{C}$ to suppress thermal noise. Owing to these characteristics, the HS output serves as a reliable ground truth. For practical deployment, we use a FUJINON HF12.5HA-1S lens with a $12.5\,\mathrm{mm}$ focal length, supporting up to a $2/3$-inch sensor. Images are captured at $800 \times 800$ resolution, corresponding to an effective imaging area of approximately $7.2\,\mathrm{mm} \times 7.2\,\mathrm{mm}$. With a 1:1 relay lens, this yields a horizontal and vertical field of view of $26.1^\circ$ and $19.5^\circ$, respectively, corresponding to $4.6\,\mathrm{m} \times 3.4\,\mathrm{m}$ at a distance of $10\,\mathrm{m}$, which is suitable for our task.

\subsection{Synthetic Process of GOPRO-BR}
~\figref{synthetic_process} illustrates our synthesizing method. Depicted as in ~\cite{liu2020deep,tao2018scale}, RS videos are generated by sequentially copying a row from corresponding high-speed frames within exposure time $\tau$, and blur observations are synthesized through averaging them. We strictly follow the constraints in \figref{synthetic_process} to ensure RS and blur views are aligned in frame level and capture identical content of the scene. Specially, We operate central crop of $512$ to each frames, and set $T = \tau = 512$.

\subsection{Training Details}

Each training and validation sample consists of an input pair $(B, R)$ and a corresponding 9-frame ground truth video clip $\textbf{G}$. The model is optimized using the Adam optimizer over 800 epochs. The learning rate starts at $10^{-4}$ and is gradually reduced to $10^{-6}$ following a cosine annealing schedule. The balance parameter $\lambda_d$ for the loss function is empirically fixed at $10^{-4}$. To enhance the diversity of training data, samples are first cropped to a resolution of $512\times512$, followed by random horizontal flips and channel reversal. All experiments are conducted on four NVIDIA Tesla V100 GPUs using a batch size of 8.
We evaluate the performance of models using standard metrics (PSNR, SSIM and LPIPS). Higher PSNR/SSIM or lower LPIPS suggests better performance.

\section{Additional Results and Discussions}

\subsection{Qualitative Visualization of Bidirectional Flows}
In the main paper, we discuss the role of bidirectional displacement fields within the SAA module for establishing cross-view correspondences between the blur and RS inputs. In this section, we provide additional qualitative visualizations (\figref{biflow_vis}) to further illustrate their behavior under different motion conditions. These examples offer more insight into how the bidirectional flows contribute to robust cross-view feature alignment.

\begin{table}[!h]
	\centering
	\caption{\textbf{Quantitative comparison with RAFT-based flow initialization.} Incorporating RAFT as an external optical flow prior degrades performance compared with our default design.}
	\label{tab:raft_initialization}
	\resizebox{1\linewidth}{!}
	{
		\begin{tabular}{rccccccccc}
			\toprule
			\multirow{2}{*}{Method} & \multicolumn{3}{c}{$\times$3} & \multicolumn{3}{c}{$\times$5} &  \multicolumn{3}{c}{$\times$9} \\ \cmidrule(lr){2-4}\cmidrule(lr){5-7}\cmidrule(lr){8-10}
			
			& PSNR     & SSIM     & LPIPS    & PSNR    & SSIM    & LPIPS & PSNR    & SSIM    & LPIPS   \\ \midrule
			
			Ours-raft & 28.44  & 0.8655 & 0.1046  &  28.63 &  0.8708 & 0.1010 & 28.74 & 0.8735 & 0.0990 \\
			Ours &30.97 & 0.9073 & 0.0695  & 31.19  &0.9105  &  0.0675 & 31.31 & 0.9122 & 0.0671  \\ \bottomrule
	\end{tabular}	}
\end{table}

\subsection{Optical Flow Initialization and External Priors}
We further investigate the impact of optical flow initialization and the use of external optical flow estimators.

In our default design, the proposed Motion Interpretation Blocks (MIBs) do not rely on any off-the-shelf optical flow estimator (e.g., RAFT or PWC-Net), nor do they adopt zero initialization. Instead, following the iterative refinement strategy in RIFE, the first block (MIB$_1$) directly predicts an initial flow from image features, and subsequent blocks progressively refine it. This design avoids introducing external flow priors or poorly estimated initial flows that may propagate errors across iterations.

To validate this design choice, we conduct additional experiments by incorporating RAFT-based flow initialization. As shown in~\tabref{raft_initialization}, introducing RAFT consistently degrades performance compared to our default design. We observe that the model tends to converge prematurely to suboptimal local minima when initialized with such external flow estimates.

We attribute this behavior to two main factors. First, there exists a significant domain gap between general-purpose optical flow estimation and our task. Second, our inputs involve severe motion degradations (e.g., blur and rolling shutter), whereas RAFT is trained on clean image pairs. As a result, the initialized flow is often biased and unreliable, which misleads subsequent refinement and harms the overall learning process.

\subsection{Upper Bound of Motion Speed}

Regarding the upper bound of motion speed, it is inherently related to the exposure time. Since RS images are captured with short row-wise exposure, the induced blur is typically limited even under fast motion. Even for very high-speed motion (e.g., fast-moving trains), the resulting blur in the RS view remains moderate in our setting, allowing it to retain useful structural cues for reconstruction.

For more extreme cases, the effect can be further mitigated in practice by reducing the exposure time, which effectively extends the applicable motion range. Although shorter exposure may introduce additional noise, especially under low-light conditions, our model has demonstrated good robustness to such degradations, as validated in Section 5.3. This suggests that the proposed framework can maintain stable performance across a wide range of motion conditions.

\subsection{Analysis of Hybrid Sensing Combinations}

Our design choice is not based merely on novelty, but on how well the sensing combination matches the two dominant ambiguities in motion degradation inversion. As discussed in the paper, blur observations are formed by temporally integrating latent frames and therefore preserve more complete global scene context, while RS observations encode row-wise temporal ordering but suffer from incomplete global content and initial-state ambiguity. This makes the RS+Blur pairing particularly well aligned with our task: the RS view helps resolve the temporal-order ambiguity of blur decomposition, whereas the blur view helps disambiguate the initial state in RS temporal super-resolution.

By comparison, RS+reversed RS (e.g., IFED) remains a strong and practical dual-frame configuration, but both views are still RS observations. Hence, its complementarity mainly arises from opposite scan directions within the same degradation family, rather than from combining temporally ordered observation with globally integrated content. In this sense, it alleviates RS ambiguity, but does not provide the same level of holistic scene context as a blur view.

RS+event (e.g., EvUnroll) offers very high temporal resolution and is effective for bridging spatio-temporal gaps. However, event streams primarily capture brightness changes rather than complete appearance/content, so they are less direct than blur observations in providing globally integrated scene information. Similarly, blur+event (e.g., EBFI) combines global intensity cues with high-temporal-resolution events, but the temporal information from events is not embedded into image coordinates as explicitly as the row-wise temporal structure in RS images, which is particularly suitable for latent-frame reasoning. Our temporal positional encoding is designed exactly to exploit this property of RS observations.

From a system perspective, RS+Blur and RS+reversed RS also enjoy a practical advantage in that they can be built using two conventional frame-based cameras, which suggests simpler hardware integration and broader deployability than event-assisted configurations. In contrast, RS+event and blur+event require heterogeneous sensing and a more complex acquisition pipeline. This practical consideration is also one reason why we regard RS+Blur as an attractive configuration beyond its reconstruction performance. The latter is confirmed in Section 5.2, where our method consistently outperforms IFED, EvUnroll, and EBFI on both synthetic and real datasets. 

Overall, among several possible hybrid sensing combinations, RS+Blur is particularly attractive because it simultaneously provides (1) explicit temporal ordering from RS, (2) globally integrated scene context from blur, and (3) a practical dual-frame-camera implementation. This principled complementarity is further validated by the experimental results in Section 5.1 and 5.2.

    \begin{IEEEbiography}[{\includegraphics[width=1in,height=1.25in,clip,keepaspectratio]{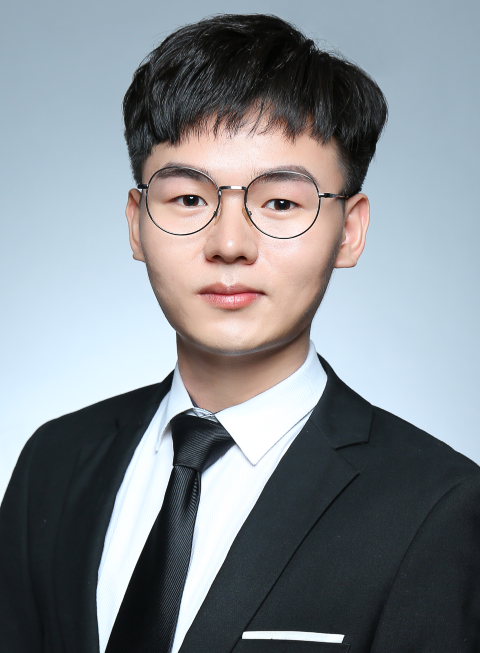}}]{Xiang Ji} received the Ph.D. degree in information science and technology from the University of Tokyo, Tokyo, Japan, in 2024. He was a research associate at the School of Computer Science and Engineering, Nanyang Technological University, Singapore, in 2021, and a research fellow  with The University of Tokyo, Japan, in 2024. He is currently a Project Assistant Professor with the Next Generation Artificial Intelligence Research Center, Graduate School of Information Science and Technology, The University of Tokyo. His research interests include physics-based vision, computational photography and computational imaging.
	\end{IEEEbiography}
    
	\begin{IEEEbiography}[{\includegraphics[width=1in,height=1.25in,clip,keepaspectratio]{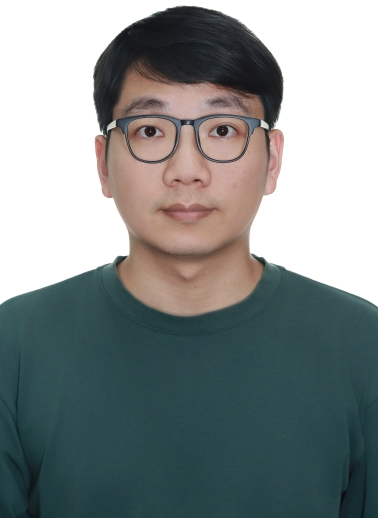}}]{Guixu Lin}
		is currently a Ph.D. student at the Graduate School of Information Science and Technology, The University of Tokyo, Japan. He received the bachelor's degree from the South China University of Technology, China, and the master's degree from The University of Tokyo, Japan. His research interests focus on video generation and event camera-based vision.
	\end{IEEEbiography}

    \begin{IEEEbiography}[{\includegraphics[width=1in,height=1.25in,clip,keepaspectratio]{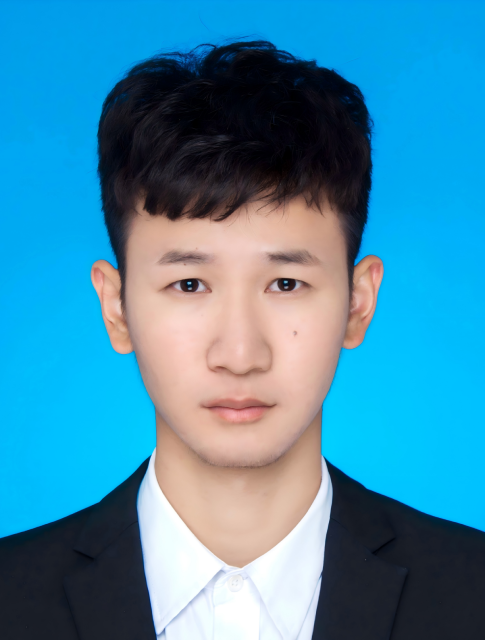}}]{Zhengwei Yin}
		is currently a Ph.D. student at the Graduate School of Information Science and Technology, The University of Tokyo, Japan. He received the bachelor's degree from the School of Geodesy and Geomatics, Wuhan University, China and the master's degree from the School of Software Engineering, University of Science and Technology of China, China. His research interests focus on image processing, large multi-modal model and their applications.
	\end{IEEEbiography}

	\begin{IEEEbiography}[{\includegraphics[width=1in,height=1.25in,clip,keepaspectratio]{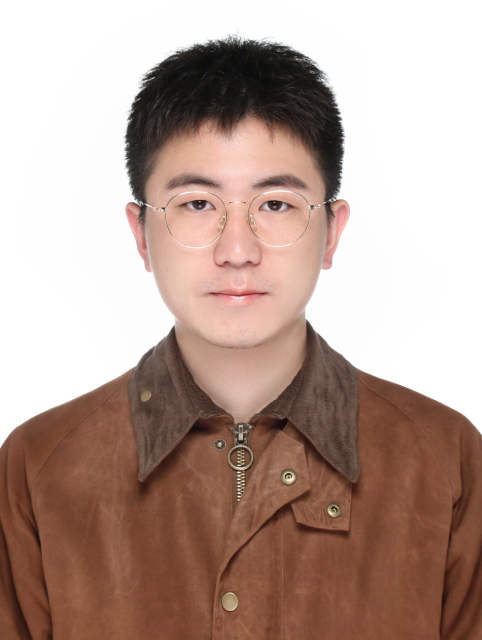}}]{Jiancheng Zhao} is currently a Ph.D. Student at the Department of Information Science and Technology, The University of Tokyo, Tokyo, Japan. He received the M.S. and B.S. degree from The University of Tokyo and University of Electronic Science and Technology of China. His research interests lie in learned image/video coding, Generative AI.
	\end{IEEEbiography}

    \begin{IEEEbiography}[{\includegraphics[width=1in,height=1.25in,clip,keepaspectratio]{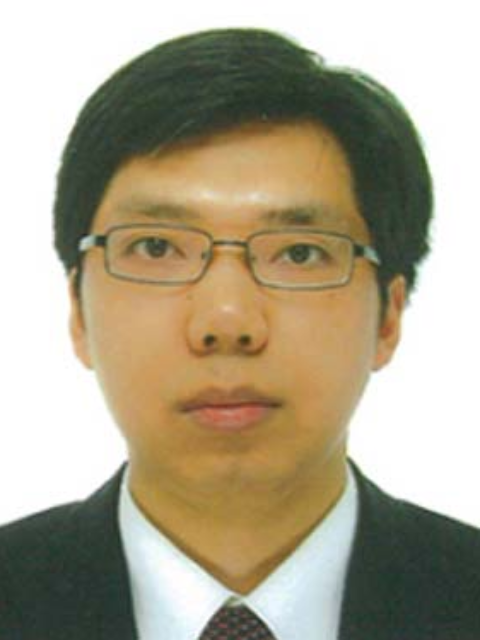}}]{Yinqiang Zheng}
    	received his doctoral degree of engineering from the Department of Mechanical and Control Engineering, Tokyo Institute of Technology, Tokyo, Japan, in 2013. He is currently a full professor in the Next Generation Artificial Intelligence Research Center, The University of Tokyo, Japan, leading the Optical Sensing and Camera System Laboratory (OSCARS Lab). He is concentrating on AI $\times$ Optical Imaging, with more than 100 research papers surrounding the novel paradigms of ‘Optics for Better AI’ and ‘AI for Best Optics’. He has served as area chair for CVPR, ICCV, ICML, ICLR, NeurIPS, MM, 3DV, ACCV, ISAIR, DICTA and MVA. He is a foreign fellow of the Engineering Academy of Japan, and the recipient of the Konica Minolta Image Science Award and Funai Academic Award. 
    \end{IEEEbiography}

\end{document}